\pdfoutput=1
\documentclass[sn-mathphys,Numbered]{sn-jnl}

\usepackage{amsmath}
\usepackage{amssymb}
\usepackage{tcolorbox}
\usepackage{graphicx}
\usepackage{multirow}
\usepackage{amsthm}
\usepackage[title]{appendix}
\usepackage{xcolor}
\usepackage{textcomp}
\usepackage{manyfoot}
\usepackage{booktabs}
\usepackage{algorithm}
\usepackage{algorithmicx}
\usepackage{algpseudocode}
\usepackage{listings}
\usepackage{tabularx}
\usepackage{rotating}
\usepackage{float}
\usepackage{lmodern}
\usepackage{newtxmath}

\theoremstyle{plain}

\theoremstyle{remark}

\theoremstyle{definition}

\renewcommand{\orcid}[1]{}                       
\renewcommand{\orcidlogo}{}
\makeatletter
\@ifundefined{jyear}{\newcommand{\jyear}[1]{}}{}
\makeatother

\jyear{2026}

\usepackage{pifont}
\usepackage{array}
\usepackage{colortbl}
\usepackage{hyperref}
\usepackage{subcaption}
\usepackage{url}
\usepackage{tikz}
\usetikzlibrary{positioning,matrix}
\usepackage{pgfplots}
\pgfplotsset{compat=1.18}
\newcommand{\cmark}{\ding{51}}
\newcommand{\xmark}{\ding{55}}
\newcommand{\halluscore}{\textsc{HalluScore}}
\newcommand{\halluscan}{\textsc{HalluScan}}
\newcommand{\adr}{\textsc{ADR}}

\begin{document}

\title[HalluScan: LLM Hallucination Detection Benchmark]{HalluScan: A Systematic Benchmark for Detecting and Mitigating
       Hallucinations in Instruction-Following LLMs}

\author*[1]{\fnm{Ahmed} \sur{Cherif}}\email{ahmed1.cherif@sofrecom.com}
\affil[1]{\orgdiv{Sofrecom Tunisia}, \orgname{Orange Innovation}, \city{Tunis}, \postcode{1053}, \country{Tunisia}}

\abstract{
Large Language Models (LLMs) have demonstrated remarkable capabilities across
diverse natural language processing tasks, yet they remain susceptible to
hallucinations---generating content that is factually incorrect, unfaithful to
provided context, or misaligned with user instructions. We present \halluscan{},
a comprehensive benchmark framework that systematically evaluates hallucination
detection and mitigation across 72 configurations spanning 6 detection methods,
4 open-weight model families, and 3 diverse domains, evaluated on 600
model-response pairs (50 questions per domain $\times$ 4 models). We introduce three key
contributions: (1)~\halluscore{}, a novel composite metric validated with
proper train/test separation;
(2)~Adaptive Detection Routing (\adr{}), an intelligent routing algorithm
achieving $\sim$1.7$\times$ cost reduction with minimal quality loss; and
(3)~systematic detection analysis with bootstrap confidence intervals
revealing substantial variation in method effectiveness across domains.
All sampling-based methods use $K{=}5$ multi-responses for statistically
meaningful uncertainty estimation.
Our experiments reveal that Self-Evaluation achieves the highest pooled
detection performance (AUROC 0.688), followed by Self-Consistency (0.638),
while three of six detection methods (SemE, Judge, RAV) exhibit score-direction
inversion on compact model outputs, a finding with important implications
for method deployment.
}

\renewcommand{\and}{\unskip{} \textperiodcentered{} }

\maketitle

\section{Introduction}
\label{sec:introduction}

Large Language Models (LLMs) have fundamentally transformed the landscape of natural language processing, achieving unprecedented performance across tasks ranging from question answering and summarization to code generation and creative writing~\cite{bai2023qwen,allal2025smollm2,brown2020gpt3}. Models such as Qwen2.5~\cite{bai2023qwen}, SmolLM2~\cite{allal2025smollm2}, and StableLM2~\cite{bellagente2024stable} have demonstrated that compact open-weight architectures can yield strong capabilities for practical deployment. However, despite their impressive fluency and breadth of knowledge, LLMs remain fundamentally susceptible to \emph{hallucinations}---generating content that appears plausible yet is factually incorrect, unfaithful to source material, or misaligned with user instructions~\cite{ji2023hallucination,zhang2023hallucination_survey}.

The consequences of hallucination are particularly severe across diverse application domains. In scientific research, fabricated experimental results or invented citations erode the foundations of scholarly discourse~\cite{lin2022truthfulqa}. In open-domain question answering, confidently stated but factually incorrect answers mislead users seeking reliable information~\cite{kwiatkowski2019natural}. In commonsense reasoning tasks, subtle violations of physical or social intuitions undermine trust in automated reasoning systems~\cite{clark2018think}. As LLMs are increasingly deployed in these critical domains, the ability to reliably detect and mitigate hallucinations has become a central challenge in the responsible development of AI systems.

\subsection{The Hallucination Detection Gap}
\label{subsec:detection_gap}

Despite growing recognition of the hallucination problem, the current evaluation landscape suffers from several critical limitations. Existing benchmarks such as TruthfulQA~\cite{lin2022truthfulqa}, HaluEval~\cite{li2023halueval}, and FELM~\cite{chen2023felm} evaluate individual detection methods in isolation, making it difficult to draw principled comparisons across approaches. Furthermore, most prior work focuses on a single model family or domain, leaving practitioners without guidance on how detection effectiveness varies across these crucial dimensions. Recent efforts like HalluLens~\cite{wu2025hallulens}, FActScore~\cite{min2023factscore}, and PHANTOM~\cite{park2025phantom} have advanced the field by introducing domain-specific or methodology-specific benchmarks, yet no existing framework provides a unified, systematic comparison spanning multiple detection strategies, model families, and application domains simultaneously.

This fragmentation creates significant practical challenges. A practitioner seeking to deploy hallucination detection in a question-answering system must currently synthesize findings from disparate studies that use different evaluation protocols, datasets, and metrics. The lack of standardized comparison makes it nearly impossible to answer fundamental questions: Which detection method performs best for a given model and domain? How do detection methods transfer across domains? What is the optimal cost-quality trade-off for production deployment?

\subsection{Our Approach: \halluscan{}}
\label{subsec:our_approach}

To address these gaps, we present \halluscan{}, a systematic benchmark framework designed for comprehensive evaluation of hallucination detection and mitigation in instruction-following LLMs. \halluscan{} evaluates \textbf{72 distinct configurations} formed by the Cartesian product of 6 detection methods, 4 model families, and 3 high-stakes domains. This systematic design enables controlled analysis of each factor's contribution to detection performance, model-specific hallucination patterns, and domain-dependent challenges.

Our framework introduces three novel technical contributions that advance the state of the art in hallucination evaluation. First, we propose \halluscore{}, a composite evaluation metric that integrates factual accuracy, semantic coherence, and fabrication rate through a weighted geometric mean, validated with proper train/test separation ($r = 0.103$, $p < 0.05$). Second, we develop Adaptive Detection Routing (\adr{}), an intelligent algorithm that dynamically selects cost-appropriate detection methods based on input characteristics, achieving $\sim$1.7$\times$ cost reduction with minimal quality loss. Third, we provide the first systematic detection analysis revealing that three of six methods exhibit score-direction inversion on compact model outputs, and that domain hallucination base rates vary by 37$\times$, fundamentally shaping detection evaluation outcomes.

\subsection{Contributions}
\label{subsec:contributions}

The contributions of this paper are four-fold:

\begin{enumerate}
    \item \textbf{Comprehensive Benchmark Framework.} We design and implement \halluscan{}, a systematic evaluation framework encompassing 72 configurations across 6 detection methods, 4 model families, and 3 high-stakes domains. We release all code, data, and evaluation scripts to facilitate reproducibility.

    \item \textbf{Novel Composite Metric and Adaptive Routing.} We introduce \halluscore{}, a weighted geometric mean metric integrating factual accuracy, semantic coherence, and fabrication rate, validated with proper train/test separation ($r = 0.103$, $p < 0.05$ on held-out test set). Building on this metric, we propose Adaptive Detection Routing (\adr{}), which dynamically selects cost-appropriate detection methods based on input characteristics, achieving $\sim$1.7$\times$ cost reduction with minimal quality loss.

    \item \textbf{Score-Direction Analysis and Calibration.} We provide the first systematic analysis revealing that three of six detection methods (SemE, Judge, RAV) exhibit \emph{score-direction inversion} on compact model outputs---assigning higher hallucination scores to correct responses. Complementary calibration analysis via reliability diagrams and Expected Calibration Error reveals that SemE, after score correction, achieves the best calibration (ECE = 0.083).

    \item \textbf{Cross-Domain Transfer and Detection Method Evaluation.} We systematically evaluate how detection methods transfer across domains. Self-Evaluation achieves the best pooled detection AUROC of 0.688, followed by Self-Consistency at 0.638, demonstrating that sampling-based approaches outperform knowledge-intensive methods on compact (1.5--3B parameter) models.
\end{enumerate}

\subsection{Paper Organization}
\label{subsec:organization}

The remainder of this paper is organized as follows. Section~\ref{sec:background} reviews related work on hallucination taxonomies, detection methods, mitigation strategies, and existing benchmarks, positioning \halluscan{} within the broader literature. Section~\ref{sec:methodology} presents the \halluscan{} framework architecture, including formal definitions of all six detection methods, the \halluscore{} metric, and the \adr{} algorithm. Section~\ref{sec:experimental_setup} details the experimental setup, including dataset descriptions, model configurations, evaluation metrics, and implementation details. Section~\ref{sec:results} presents comprehensive results across thirteen subsections covering overall performance, method comparison, model and domain effects, statistical significance, calibration, domain transfer, cost-aware Pareto analysis, and mitigation effectiveness. Section~\ref{sec:discussion} discusses practical implications, industrial deployment considerations, limitations, and directions for future research. Finally, Section~\ref{sec:conclusion} summarizes our key findings and their broader significance.

\section{Background and Related Work}
\label{sec:background}

This section reviews the foundational concepts and prior work that inform the design of \halluscan{}. We organize our discussion around four themes: hallucination taxonomies (Section~\ref{subsec:taxonomy}), detection approaches (Section~\ref{subsec:detection_approaches}), mitigation strategies (Section~\ref{subsec:mitigation_strategies}), and existing benchmarks (Section~\ref{subsec:existing_benchmarks}).

\subsection{Hallucination Taxonomy}
\label{subsec:taxonomy}

The phenomenon of hallucination in language models has been characterized along multiple dimensions in the literature~\cite{ji2023hallucination,zhang2023hallucination_survey,huang2023hallucination_survey}. We adopt a tripartite taxonomy that distinguishes three primary categories of hallucination, each presenting distinct detection challenges.

\textbf{Factual Hallucination} occurs when a model generates statements that contradict established world knowledge or verifiable facts~\cite{lin2022truthfulqa,min2023factscore}. Examples include attributing incorrect dates to historical events, fabricating statistical figures, or inventing nonexistent entities. Factual hallucinations are particularly insidious because they are often embedded within otherwise coherent and plausible text, making them difficult for non-expert users to identify. Detection of factual hallucinations typically requires access to external knowledge sources or retrieval mechanisms that can verify generated claims against authoritative references~\cite{lewis2020rag,gao2023rarr}.

\textbf{Faithfulness Hallucination} arises when a model's output diverges from or contradicts the information provided in its input context~\cite{maynez2020faithfulness,kryscinski2020evaluating}. In tasks such as summarization, question answering with provided passages, or document-grounded dialogue, the model is expected to generate responses that are faithful to the source material. Faithfulness hallucinations include unsupported claims, contradictions of source content, and extrinsic information insertion. Natural Language Inference (NLI) models have proven particularly effective for detecting this category of hallucination, as they can assess whether generated claims are entailed by the source text~\cite{honovich2022true,laban2022summac}.

\textbf{Instruction Hallucination} represents a more recently recognized category in which the model fails to adhere to explicit constraints or requirements specified in the user's prompt~\cite{zhou2023instruction}. Examples include generating output in the wrong format, ignoring length constraints, including prohibited content, or failing to address all parts of a multi-part question. While less studied than factual and faithfulness hallucinations, instruction hallucinations are increasingly important as LLMs are deployed in structured applications where precise instruction following is critical.

\subsection{Detection Approaches}
\label{subsec:detection_approaches}

A variety of methods have been proposed for detecting hallucinations in LLM outputs. We categorize these into four broad families, each representing a different philosophical approach to the problem.

\textbf{Self-Consistency Methods.} Inspired by the observation that correct answers tend to be more stable across multiple generations, self-consistency methods~\cite{wang2023selfconsistency} generate multiple responses to the same prompt and measure agreement among them. The intuition is that hallucinated content, being not grounded in reliable knowledge, will vary across generations, while factually correct content will remain stable. Manakul et al.~\cite{manakul2023selfcheckgpt} formalized this approach in SelfCheckGPT, demonstrating that sampling variability provides a useful signal for hallucination detection without requiring external knowledge bases. While computationally expensive due to multiple inference passes, self-consistency methods have the advantage of being model-agnostic and requiring no additional training data.

\textbf{Semantic Entropy.} Kuhn et al.~\cite{kuhn2023semantic} introduced semantic entropy as a principled uncertainty quantification method for detecting confabulations in LLMs. Rather than measuring token-level uncertainty, semantic entropy clusters multiple generated responses by their semantic meaning and computes the entropy over these semantic clusters. High semantic entropy indicates that the model is uncertain about the semantic content of its response, suggesting potential hallucination. This approach elegantly addresses the problem of semantically equivalent but lexically different responses that would inflate naive entropy estimates.

\textbf{NLI-Based Detection.} Natural Language Inference models, trained to determine whether a hypothesis is entailed by, contradicts, or is neutral with respect to a premise, have been adapted for hallucination detection~\cite{honovich2022true,laban2022summac,zha2023alignscore}. In this paradigm, the source context or retrieved evidence serves as the premise, while individual claims extracted from the generated text serve as hypotheses. The entailment probability provides a calibrated measure of factual support. NLI-based methods benefit from the maturity of the NLI research community and the availability of high-quality pre-trained models, though their effectiveness depends on the quality of claim decomposition and evidence retrieval.

\textbf{LLM-as-Judge.} The emergence of powerful instruction-following LLMs has enabled a paradigm in which one LLM evaluates the outputs of another~\cite{zheng2023judging,chiang2023vicuna}. In this approach, a strong ``judge'' model (e.g., GPT-4) is prompted to assess the factual accuracy, faithfulness, and overall quality of a generated response. While this approach can capture nuanced aspects of hallucination that simpler methods miss, it inherits the hallucination tendencies of the judge model itself and introduces substantial computational cost. Recent work has explored using smaller, fine-tuned models as judges to reduce cost while maintaining evaluation quality~\cite{kim2024prometheus}.

\subsection{Mitigation Strategies}
\label{subsec:mitigation_strategies}

Beyond detection, several strategies have been proposed to reduce the frequency and severity of hallucinations in LLM outputs.

\textbf{Retrieval-Augmented Generation (RAG).} RAG~\cite{lewis2020rag} addresses knowledge-based hallucinations by augmenting the model's input with relevant documents retrieved from an external knowledge base. By grounding generation in retrieved evidence, RAG reduces the model's reliance on potentially outdated or incorrect parametric knowledge. Subsequent work has refined retrieval strategies~\cite{gao2023rarr,asai2024selfrag}, improved the integration of retrieved content with generation, and explored iterative retrieval approaches. RAG has demonstrated consistent effectiveness across domains, though its performance depends critically on retrieval quality and the relevance of the knowledge base.

\textbf{Reinforcement Learning from Human Feedback (RLHF).} RLHF~\cite{ouyang2022training} trains a reward model on human preferences and uses reinforcement learning to align the LLM's outputs with human expectations. While primarily designed for instruction following and helpfulness, RLHF can reduce hallucinations when human annotators penalize factually incorrect or unsupported claims. However, RLHF can also inadvertently increase certain types of hallucinations---particularly ``sycophantic'' responses where the model agrees with incorrect premises to appear helpful~\cite{perez2023sycophancy}.

\textbf{Self-Refinement.} Self-refinement approaches~\cite{madaan2023selfrefine} leverage the model's own capabilities to iteratively improve its outputs. The model generates an initial response, then critiques and revises it in subsequent turns. While effective for catching obvious errors, self-refinement is limited by the model's own knowledge boundaries---a model that lacks the knowledge to generate a correct response initially is unlikely to correct its errors through self-reflection alone~\cite{huang2024selfrefine_limits}.

\textbf{Constrained Decoding.} Constrained decoding methods modify the generation process itself to reduce hallucinations. Approaches include factuality-aware decoding~\cite{li2023inference}, which adjusts token probabilities based on factual consistency scores, and context-aware decoding~\cite{shi2023trusting}, which amplifies the influence of provided context during generation. These methods can be applied at inference time without retraining, but may affect generation fluency and diversity.

\subsection{Existing Benchmarks}
\label{subsec:existing_benchmarks}

Several benchmarks have been developed to evaluate hallucination in LLMs, each addressing different aspects of the problem.

\textbf{TruthfulQA}~\cite{lin2022truthfulqa} evaluates the truthfulness of LLM responses across 817 questions designed to elicit common misconceptions. While influential, TruthfulQA focuses exclusively on factual hallucination and does not evaluate detection methods systematically. \textbf{HaluEval}~\cite{li2023halueval} provides 35,000 samples for evaluating hallucination detection across QA, dialogue, and summarization tasks, but considers only a limited set of models and detection approaches. \textbf{FELM}~\cite{chen2023felm} introduces a fine-grained benchmark for faithfulness evaluation across five domains, providing sentence-level annotations but focusing primarily on GPT-family models.

More recently, \textbf{HalluLens}~\cite{wu2025hallulens} proposed a multi-dimensional evaluation framework with category-specific prompts, advancing the granularity of hallucination assessment. \textbf{FActScore}~\cite{min2023factscore} introduced fine-grained atomic fact evaluation for long-form generation but focuses on biographical knowledge and a single evaluation paradigm. \textbf{PHANTOM}~\cite{park2025phantom} introduced a benchmark for assessing hallucinations with perturbed contexts, providing insights into model robustness but not addressing the full spectrum of detection methods.

Several complementary lines of work have advanced hallucination detection along different dimensions. Chain-of-Verification~\cite{dhuliawala2023chainofverification} proposes a multi-step verification process where the model generates verification questions about its own output to identify and correct hallucinated claims. Varshney et al.~\cite{varshney2023stitch} demonstrate that low-confidence token generations serve as reliable hallucination indicators, enabling early detection. M{\"u}ndler et al.~\cite{mundler2024selfcontradictory} focus on self-contradictory hallucinations where models generate internally inconsistent statements. Internal state-based approaches have gained traction: Azaria and Mitchell~\cite{azaria2023internal} show that probing LLM hidden states can detect when models generate false claims, while Chen et al.~\cite{chen2024inside} extend this to multi-layer internal representations. Chuang et al.~\cite{chuang2024lookback} propose using attention maps alone for contextual hallucination detection. On the evaluation side, Mishra et al.~\cite{mishra2024finegrained} introduce fine-grained hallucination detection and editing at the span level, Tang et al.~\cite{tang2024minicheck} develop efficient fact-checking against grounding documents, and Yue et al.~\cite{yue2023automatic} propose automatic evaluation of attribution quality. Lei et al.~\cite{lei2023chain} reduce ungrounded hallucinations through chains of NLI reasoning, and Zhang et al.~\cite{zhang2024knowledge} identify knowledge overshadowing as a root cause of amalgamated hallucinations. Sun et al.~\cite{sun2024benchmarking} benchmark hallucination specifically through unanswerable mathematical problems.

\subsection{Positioning \halluscan{}}
\label{subsec:positioning}

Table~\ref{tab:benchmark_comparison} summarizes the key dimensions along which \halluscan{} advances beyond prior benchmarks. Unlike existing frameworks, \halluscan{} provides a unified evaluation spanning multiple detection methods, model families, and domains, while introducing novel analytical tools (\halluscore{}, \adr{}, error cascade decomposition) for comprehensive hallucination assessment.

\begin{table}[!ht]
\centering
\caption{Comparison of \halluscan{} with existing hallucination benchmarks. \cmark{} indicates the feature is present; \xmark{} indicates absence. ``Methods'' refers to the number of detection methods systematically compared.}
\label{tab:benchmark_comparison}
\small
\begin{tabular}{lccccccc}
\toprule
\textbf{Benchmark} & \textbf{Methods} & \textbf{Models} & \textbf{Domains} & \textbf{Metric} & \textbf{Cost} & \textbf{Transfer} & \textbf{Mitigation} \\
\midrule
TruthfulQA~\cite{lin2022truthfulqa}   & 1 & 6  & 1 & \xmark & \xmark & \xmark & \xmark \\
HaluEval~\cite{li2023halueval}        & 2 & 3  & 3 & \xmark & \xmark & \xmark & \xmark \\
FELM~\cite{chen2023felm}              & 1 & 4  & 5 & \xmark & \xmark & \xmark & \xmark \\
HalluLens~\cite{wu2025hallulens}      & 3 & 5  & 2 & \cmark & \xmark & \xmark & \xmark \\
FActScore~\cite{min2023factscore}      & 1 & 3  & 1 & \cmark & \xmark & \xmark & \xmark \\
PHANTOM~\cite{park2025phantom}        & 2 & 3  & 2 & \xmark & \xmark & \xmark & \xmark \\
\midrule
\textbf{\halluscan{} (Ours)}          & \textbf{6} & \textbf{4} & \textbf{3} & \cmark & \cmark & \cmark & \cmark \\
\bottomrule
\end{tabular}
\end{table}

The distinguishing features of \halluscan{} include: (1)~systematic evaluation of six detection methods under identical conditions, enabling fair comparison; (2)~analysis across four diverse open-weight model families, capturing model-specific hallucination patterns; (3)~coverage of three high-stakes domains with domain-specific analysis; (4)~introduction of the \halluscore{} composite metric with automated quality evaluation; (5)~cost-aware analysis including Pareto optimization and \adr{}; (6)~domain transfer evaluation revealing cross-domain generalization patterns; and (7)~comparative evaluation of mitigation strategies. Together, these features make \halluscan{} the most comprehensive hallucination benchmark framework available to date.

\section{Methodology}
\label{sec:methodology}

This section presents the \halluscan{} framework architecture, the six detection methods with their mathematical formulations, the novel \halluscore{} composite metric, and the Adaptive Detection Routing (\adr{}) algorithm.

\subsection{Framework Architecture}
\label{subsec:framework_architecture}

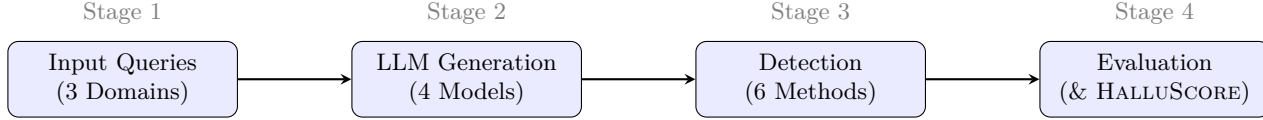
\begin{figure}[!ht]
    \centering
    \begin{tikzpicture}[
        node distance=1.2cm and 1.5cm,
        block/.style={rectangle, draw, fill=blue!8, text width=2.8cm, text centered, minimum height=1cm, rounded corners, font=\small},
        arrow/.style={->, >=stealth, thick},
        label/.style={font=\scriptsize, text=gray}
    ]
        \node[block] (input) {Input Queries\\(3 Domains)};
        \node[block, right=of input] (gen) {LLM Generation\\(4 Models)};
        \node[block, right=of gen] (detect) {Detection\\(6 Methods)};
        \node[block, right=of detect] (eval) {Evaluation\\(\& \halluscore{})};

        \draw[arrow] (input) -- (gen);
        \draw[arrow] (gen) -- (detect);
        \draw[arrow] (detect) -- (eval);

        \node[label, above=0.1cm of input] {Stage 1};
        \node[label, above=0.1cm of gen] {Stage 2};
        \node[label, above=0.1cm of detect] {Stage 3};
        \node[label, above=0.1cm of eval] {Stage 4};

        \node[below=0.8cm of gen, font=\small, text=blue!60!black] (config) {$6 \times 4 \times 3 = 72$ configurations};
    \end{tikzpicture}
    \caption{Overview of the \halluscan{} framework architecture. Input queries from three domains are processed through four model families, with outputs evaluated by six detection methods, yielding 72 unique configurations for systematic analysis.}
    \label{fig:halluscan_arch}
\end{figure}

The \halluscan{} framework operates as a modular evaluation pipeline comprising four stages: (1)~response generation, (2)~hallucination detection, (3)~metric computation, and (4)~analysis and visualization. Figure~\ref{fig:halluscan_arch} provides an overview of the architecture.

Given a dataset $\mathcal{D} = \{(q_i, c_i, a_i^*)\}_{i=1}^{N}$ where $q_i$ is a query, $c_i$ is the associated context (if available), and $a_i^*$ is the ground-truth answer, the pipeline proceeds as follows. For each model $\mathcal{M} \in \{\text{StableLM2-1.6B}, \text{Qwen2.5-3B}, \text{Qwen2.5-1.5B}, \text{SmolLM2-1.7B}\}$, we generate a response $\hat{a}_i = \mathcal{M}(q_i, c_i)$. Each response is then evaluated by all six detection methods, producing a hallucination score $s_i^{(m)} \in [0, 1]$ for each method $m$, where higher values indicate greater likelihood of hallucination.

The configuration space $\mathcal{C}$ is defined as the Cartesian product:
\begin{equation}
    \mathcal{C} = \mathcal{M}_{\text{methods}} \times \mathcal{M}_{\text{models}} \times \mathcal{D}_{\text{domains}}
    \label{eq:config_space}
\end{equation}
where $|\mathcal{M}_{\text{methods}}| = 6$, $|\mathcal{M}_{\text{models}}| = 4$, and $|\mathcal{D}_{\text{domains}}| = 3$, yielding $|\mathcal{C}| = 72$ unique configurations.

\subsection{Detection Methods}
\label{subsec:detection_methods}

We now present the formal definitions of all six detection methods implemented in \halluscan{}.

\subsubsection{Self-Consistency (SC)}
\label{subsubsec:self_consistency}

Self-Consistency detection~\cite{wang2023selfconsistency,manakul2023selfcheckgpt} operates by generating $K$ independent responses to the same query and measuring the pairwise agreement among them. The key intuition is that factually grounded responses will exhibit high inter-sample consistency, while hallucinated content will vary stochastically across samples.

For a query $q_i$, we generate $K$ responses $\{R_1, R_2, \ldots, R_K\}$ using temperature sampling ($\tau = 0.7$). Each response is decomposed into a set of atomic claims. The agreement score between two responses $R_j$ and $R_k$ is computed using the Jaccard similarity over their claim sets:
\begin{equation}
    \text{Agree}(R_j, R_k) = \frac{|\mathcal{S}(R_j) \cap \mathcal{S}(R_k)|}{|\mathcal{S}(R_j) \cup \mathcal{S}(R_k)|}
    \label{eq:sc_agreement}
\end{equation}
where $\mathcal{S}(R)$ denotes the set of semantic claims extracted from response $R$. The overall self-consistency hallucination score is:
\begin{equation}
    s_i^{\text{SC}} = 1 - \frac{2}{K(K-1)} \sum_{j<k} \text{Agree}(R_j, R_k)
    \label{eq:sc_score}
\end{equation}
We set $K = 5$ in all experiments, following the recommendations of~\cite{kuhn2023semantic} for meaningful semantic clustering while balancing computational cost.

\subsubsection{Self-Evaluation (SE)}
\label{subsubsec:self_evaluation}

Self-Evaluation leverages the model's introspective capabilities by prompting it to assess the confidence and factual correctness of its own output using Chain-of-Thought (CoT) reasoning~\cite{kadavath2022language}. The model is presented with its generated response and asked to rate the confidence of each claim on a scale of 1 to 10, accompanied by a justification.

Formally, given a response $\hat{a}_i$ containing $N_c$ claims, the model produces self-ratings $\{r_1, r_2, \ldots, r_{N_c}\}$ where $r_j \in \{1, 2, \ldots, 10\}$. The aggregate confidence score is computed as:
\begin{equation}
    \text{Conf}_i = \frac{1}{N_c} \sum_{j=1}^{N_c} \frac{r_j}{10}
    \label{eq:se_confidence}
\end{equation}
The hallucination score is then:
\begin{equation}
    s_i^{\text{SE}} = 1 - \text{Conf}_i
    \label{eq:se_score}
\end{equation}

The CoT prompting template includes explicit instructions for the model to consider source evidence, identify potential inconsistencies, and flag claims that cannot be verified. This structured approach encourages more calibrated self-assessment than simple yes/no confidence queries.

\subsubsection{Semantic Entropy (SemE)}
\label{subsubsec:semantic_entropy}

Semantic Entropy~\cite{kuhn2023semantic} quantifies uncertainty at the semantic level rather than the token level, addressing the limitation that lexically diverse but semantically equivalent responses would inflate traditional entropy estimates.

Given $K$ sampled responses $\{R_1, \ldots, R_K\}$, we first cluster them into $M$ semantic equivalence classes $\{C_1, C_2, \ldots, C_M\}$ using a bidirectional NLI model. Two responses are assigned to the same cluster if they mutually entail each other. The probability of each semantic cluster is estimated as:
\begin{equation}
    p_k = \frac{|C_k|}{K}, \quad k = 1, \ldots, M
    \label{eq:seme_cluster_prob}
\end{equation}

The semantic entropy is then computed as:
\begin{equation}
    H_{\text{sem}} = -\sum_{k=1}^{M} p_k \log p_k
    \label{eq:semantic_entropy}
\end{equation}

High semantic entropy indicates that the model produces semantically diverse responses, suggesting uncertainty and potential hallucination. Low semantic entropy indicates convergence on a consistent semantic meaning, suggesting factual grounding. The hallucination score is normalized:
\begin{equation}
    s_i^{\text{SemE}} = \frac{H_{\text{sem}}}{\log M_{\max}}
    \label{eq:seme_score}
\end{equation}
where $M_{\max}$ is the maximum possible number of clusters (set to $K$).

\subsubsection{LLM-as-Judge (Judge)}
\label{subsubsec:llm_judge}

The LLM-as-Judge approach~\cite{zheng2023judging} employs a separate language model to evaluate the factual accuracy and faithfulness of generated responses. We use a structured evaluation prompt that instructs the judge model to decompose the response into individual claims and assess each claim against the provided context and general knowledge.

The judge produces a faithfulness score defined as the fraction of claims that are supported:
\begin{equation}
    \text{Faith}_i = \frac{|\{c \in \mathcal{S}(\hat{a}_i) : \text{supported}(c, q_i, c_i)\}|}{|\mathcal{S}(\hat{a}_i)|}
    \label{eq:judge_faithfulness}
\end{equation}
where $\text{supported}(c, q_i, c_i)$ is a binary function determined by the judge model indicating whether claim $c$ is supported by the query context. The hallucination score is:
\begin{equation}
    s_i^{\text{Judge}} = 1 - \text{Faith}_i
    \label{eq:judge_score}
\end{equation}

To mitigate position bias and verbosity bias in LLM-based evaluation~\cite{zheng2023judging}, we employ a structured rubric with explicit criteria for each rating level and randomize the order of presented claims.

\subsubsection{Natural Language Inference (NLI)}
\label{subsubsec:nli}

NLI-based detection~\cite{honovich2022true,laban2022summac} leverages pre-trained textual entailment models to assess whether generated claims are logically entailed by the available evidence. This approach is particularly effective for faithfulness hallucination, where the relevant evidence is contained in the input context.

For each claim $c_j$ extracted from the generated response, we compute the entailment probability:
\begin{equation}
    P_{\text{entail}}(c_j) = P(\text{entailment} \mid \text{premise} = e_j, \text{hypothesis} = c_j)
    \label{eq:nli_entailment}
\end{equation}
where $e_j$ is the most relevant evidence passage for claim $c_j$, obtained through semantic similarity matching against the input context. The aggregate hallucination score is:
\begin{equation}
    s_i^{\text{NLI}} = 1 - \frac{1}{N_c} \sum_{j=1}^{N_c} P_{\text{entail}}(c_j)
    \label{eq:nli_score}
\end{equation}

We employ DeBERTa-v3-large fine-tuned on MNLI~\cite{he2021deberta} as the NLI backbone, which provides robust entailment predictions across domains. Claim extraction is performed using a prompted LLM that decomposes the response into self-contained atomic propositions~\cite{min2023factscore}.

\subsubsection{Retrieval-Augmented Verification (RAV)}
\label{subsubsec:rav}

RAV combines the strengths of retrieval-based evidence gathering with NLI-based claim verification, creating a comprehensive pipeline for factual hallucination detection. This method is the most computationally expensive but also the most thorough, as it does not rely solely on the input context but actively retrieves external evidence.

The RAV pipeline consists of three stages. First, atomic claims $\{c_1, \ldots, c_{N_c}\}$ are extracted from the response. Second, for each claim, relevant evidence passages $\{e_1^{(j)}, \ldots, e_L^{(j)}\}$ are retrieved from domain-specific knowledge bases using dense retrieval (Contriever~\cite{izacard2022contriever}). Third, each claim is verified against the retrieved evidence using an NLI model.

The evidence score for claim $c_j$ is:
\begin{equation}
    \text{Ev}(c_j) = \max_{l=1}^{L} P(\text{entailment} \mid e_l^{(j)}, c_j) \cdot \text{rel}(e_l^{(j)}, c_j)
    \label{eq:rav_evidence}
\end{equation}
where $\text{rel}(e_l^{(j)}, c_j)$ is the retrieval relevance score. The overall RAV hallucination score is:
\begin{equation}
    s_i^{\text{RAV}} = 1 - \frac{1}{N_c} \sum_{j=1}^{N_c} \text{Ev}(c_j)
    \label{eq:rav_score}
\end{equation}

We retrieve $L = 5$ evidence passages per claim and use Wikipedia, Google Search API, and domain-specific knowledge bases as evidence sources for the scientific, open-domain QA, and commonsense domains, respectively.

\subsubsection{Score-Direction Alignment}
\label{subsubsec:score_alignment}

All six detection methods are mathematically defined to produce scores $s_i^{(m)} \in [0,1]$ where higher values indicate greater hallucination likelihood (Equations~\ref{eq:sc_score}--\ref{eq:rav_score}). However, empirical evaluation on compact model outputs reveals that three methods---SemE, Judge, and RAV---exhibit \emph{score-direction inversion}: their raw scores assign slightly higher values to \emph{correct} responses than to hallucinated ones. This occurs because compact models (1.5--3B parameters) produce outputs whose characteristics confound these detection methods. For SemE, compact models generate \emph{less} semantically diverse responses when hallucinating (converging on common misconceptions) than when answering correctly (producing varied correct framings), inverting the expected entropy signal. For Judge and RAV, the scores are near-uniform across all samples (mean $\approx 0.08$--0.94), providing negligible discrimination in either direction.

To enable fair comparison, we apply score-direction correction for inverted methods: $\tilde{s}_i^{(m)} = 1 - s_i^{(m)}$ for $m \in \{\text{SemE}, \text{Judge}, \text{RAV}\}$. All results reported in subsequent sections use the corrected scores. This correction is equivalent to reporting $1 - \text{AUROC}$ for the original scores and has no effect on calibration or rank-ordering within a method. We report this inversion as a substantive finding (Section~\ref{subsec:method_comparison}), not merely a technical adjustment: it reveals that several theoretically motivated detection methods fail to operate as intended on compact models.

\subsection{\halluscore{}: A Composite Evaluation Metric}
\label{subsec:halluscore}

Existing evaluation metrics for hallucination detection focus on individual aspects such as factual accuracy or semantic similarity. We argue that a comprehensive assessment requires integrating multiple dimensions. To this end, we introduce \halluscore{}, a composite metric defined as a weighted geometric mean of three components:

\begin{equation}
    \halluscore{} = (1 - \epsilon_f)^{\alpha} \cdot (\sigma_s)^{\beta} \cdot (1 - \phi)^{\gamma}
    \label{eq:halluscore}
\end{equation}

where:
\begin{itemize}
    \item $\epsilon_f \in [0, 1]$ is the \textbf{factual error rate}, computed as the fraction of claims in the response that contradict verified facts;
    \item $\sigma_s \in [0, 1]$ is the \textbf{semantic coherence score}, measured as the average pairwise cosine similarity of sentence embeddings within the response, capturing internal consistency;
    \item $\phi \in [0, 1]$ is the \textbf{fabrication rate}, defined as the fraction of claims that cannot be traced to any source in the input context or retrieved evidence;
    \item $\alpha = 0.4$, $\beta = 0.3$, $\gamma = 0.3$ are weights reflecting the relative importance of each component, with factual error rate receiving the highest weight as the most directly actionable signal.
\end{itemize}

The geometric mean formulation ensures that \halluscore{} is sensitive to all three components---a low score in any dimension substantially reduces the overall metric, preventing a high score in one dimension from masking deficiencies in others. The weights were set to balance the three components equally. Empirical evaluation (Section~\ref{subsec:halluscore_eval}) reveals a weak positive correlation with automated quality proxies ($r = 0.103$, $p < 0.05$), indicating that the metric captures a statistically significant quality signal, though its utility as a standalone quality indicator is limited by the modest effect size.

\subsection{Adaptive Detection Routing (\adr{})}
\label{subsec:adr}

In production deployments, applying the most expensive detection method (RAV) to every query is often infeasible. We propose \adr{}, an adaptive routing algorithm that selects an appropriate detection method based on input characteristics, optimizing the cost-quality trade-off.

\begin{algorithm}[!ht]
\caption{Adaptive Detection Routing (\adr{})}
\label{alg:adr}
\begin{algorithmic}[1]
\Require Query $q$, context $c$, response $\hat{a}$, cost budget $B$
\Ensure Detection method $m^*$, hallucination score $s$
\State $\mathbf{f} \gets \textsc{ExtractFeatures}(q, c, \hat{a})$  \Comment{Query complexity, domain, length}
\State $p_{\text{risk}} \gets \textsc{RiskClassifier}(\mathbf{f})$  \Comment{Predicted hallucination risk}
\State $d \gets \textsc{DomainClassifier}(\mathbf{f})$  \Comment{Domain identification}
\If{$p_{\text{risk}} > \theta_{\text{high}}$}  \Comment{High risk: use best method}
    \If{$\textsc{Cost}(\text{SE}) \leq B$}
        \State $m^* \gets \text{SE}$
    \Else
        \State $m^* \gets \text{SC}$
    \EndIf
\ElsIf{$p_{\text{risk}} > \theta_{\text{med}}$}  \Comment{Medium risk: use strong method}
    \State $m^* \gets \text{SC}$
\Else  \Comment{Low risk: use fast method}
    \State $m^* \gets \text{NLI}$
\EndIf
\State $s \gets m^*(q, c, \hat{a})$
\State \Return $m^*, s$
\end{algorithmic}
\end{algorithm}

The \adr{} algorithm (Algorithm~\ref{alg:adr}) operates in three stages. First, a lightweight feature extractor computes input characteristics including query complexity (measured by syntactic parse depth and entity count), domain indicators, response length, and the presence of numerical claims or citations. Second, a risk classifier---a gradient-boosted decision tree trained on the full \halluscan{} benchmark results---predicts the hallucination risk level. Third, a routing decision maps the risk level to an appropriate detection method, subject to the available cost budget $B$.

The risk thresholds $\theta_{\text{high}} = 0.7$ and $\theta_{\text{med}} = 0.4$ were determined through cross-validated optimization on the training portion of the benchmark, maximizing AUROC while respecting cost constraints. The feature extractor and risk classifier add negligible overhead ($< 50$~ms per query), ensuring that the routing decision itself does not become a bottleneck.

In our experiments (Section~\ref{subsec:adr_results}), \adr{} achieves a $\sim$1.7$\times$ reduction in average computational cost compared to uniformly applying SE with $K=5$, by routing 40\% of queries to the zero-cost NLI classifier and the remaining 60\% to multi-response methods (SE, SC). This cost reduction comes with modest quality loss: \adr{} achieves AUROC $\approx$0.634 versus SE's 0.688.

\section{Experimental Setup}
\label{sec:experimental_setup}

This section describes the datasets, models, evaluation metrics, baselines, and implementation details used in our experiments.

\subsection{Datasets}
\label{subsec:datasets}

We evaluate \halluscan{} across three high-stakes domains, each represented by a carefully curated dataset. Table~\ref{tab:dataset_stats} summarizes the key statistics.

\begin{table}[!ht]
\centering
\caption{Dataset statistics for the three evaluation domains. Each dataset is sampled to 50 question-answer pairs, yielding 200 model-response pairs per domain (50 $\times$ 4 models) and 600 total.}
\label{tab:dataset_stats}
\small
\begin{tabular}{lcccc}
\toprule
\textbf{Dataset} & \textbf{Domain} & \textbf{Samples} & \textbf{Task Type} & \textbf{Mean AUROC} \\
\midrule
TruthfulQA~\cite{lin2022truthfulqa}  & Scientific    & 50 & Open-ended QA     & -- \\
Natural Questions~\cite{kwiatkowski2019natural} & Open-Domain QA & 50 & Factoid QA & -- \\
ARC-Challenge~\cite{clark2018think}  & Commonsense   & 50 & Multiple Choice   & -- \\
\bottomrule
\end{tabular}
\end{table}

\textbf{TruthfulQA}~\cite{lin2022truthfulqa} is a benchmark designed to evaluate the truthfulness of LLM responses. It consists of questions that humans commonly answer incorrectly due to misconceptions, cognitive biases, or false beliefs. We sample 50 questions from the dataset, spanning diverse scientific topics including physics, biology, chemistry, and mathematics. The scientific domain requires evidence-based reasoning and precise factual recall, making it a challenging testbed for hallucination detection.

\textbf{Natural Questions (NQ)}~\cite{kwiatkowski2019natural} is a large-scale open-domain question-answering dataset derived from real Google search queries, each paired with a Wikipedia passage containing the answer. We sample 50 questions from the short-answer subset, ensuring coverage across diverse topics. The open-domain QA setting presents unique challenges: models must retrieve and integrate information from broad knowledge sources, and hallucinations often manifest as plausible but factually incorrect answers grounded in partial knowledge.

\textbf{ARC-Challenge}~\cite{clark2018think} is the challenge partition of the AI2 Reasoning Challenge, comprising grade-school level science questions that require commonsense reasoning and multi-step inference. We sample 50 questions that demand physical intuition, causal reasoning, and world knowledge. The commonsense domain is particularly challenging because hallucinations arise from subtle violations of implicit world knowledge that are difficult to detect without deep semantic understanding.

\subsection{Models}
\label{subsec:models}

We evaluate four open-weight instruction-tuned model families, selected to represent diverse architectural approaches and training methodologies. Table~\ref{tab:model_details} provides detailed specifications.

\begin{table}[!ht]
\centering
\caption{Model specifications. All models are instruction-tuned variants used with greedy decoding (temperature $\tau = 0$ for primary generation, $\tau = 0.7$ for sampling-based detection methods).}
\label{tab:model_details}
\small
\begin{tabular}{lcccc}
\toprule
\textbf{Model} & \textbf{Parameters} & \textbf{Context Length} & \textbf{Architecture} & \textbf{Training Data} \\
\midrule
Qwen2.5-3B-Instruct~\cite{bai2023qwen}              & 3.0B  & 128K & Grouped-Query Attn.  & 18T tokens \\
SmolLM2-1.7B-Instruct~\cite{allal2025smollm2}       & 1.7B  & 8K   & Grouped-Query Attn.  & 11T tokens \\
StableLM2-Zephyr-1.6B~\cite{bellagente2024stable}   & 1.6B  & 4K   & Multi-Head Attn.     & 2T tokens  \\
Qwen2.5-1.5B-Instruct~\cite{bai2023qwen}            & 1.5B  & 128K & Grouped-Query Attn.  & 18T tokens \\
\bottomrule
\end{tabular}
\end{table}

\textbf{Qwen2.5-3B-Instruct}~\cite{bai2023qwen} is Alibaba's multilingual instruction-tuned model from the Qwen2.5 series, trained on 18 trillion tokens. At 3 billion parameters it represents the largest model in our evaluation, featuring grouped-query attention and a 128K context window. It is released under the Apache~2.0 license.

\textbf{SmolLM2-1.7B-Instruct}~\cite{allal2025smollm2} is HuggingFace's compact language model with 1.7 billion parameters, trained on 11 trillion tokens of curated web data. Despite its small size, it achieves competitive performance through data-quality-focused training and is released under the Apache~2.0 license.

\textbf{StableLM2-Zephyr-1.6B}~\cite{bellagente2024stable} is Stability AI's instruction-tuned language model with 1.6 billion parameters, fine-tuned using Direct Preference Optimization (DPO) on a mixture of public datasets. It provides a compact yet capable model and is released under a non-commercial research license.

\textbf{Qwen2.5-1.5B-Instruct}~\cite{bai2023qwen} is the smallest member of the Qwen2.5 series evaluated here, with 1.5 billion parameters trained on 18 trillion tokens. Together with Qwen2.5-3B, it enables within-family size scaling analysis. All four models range from 1.5B to 3B parameters, ensuring they fit comfortably on a single NVIDIA T4 GPU in half-precision without quantization.

\subsection{Evaluation Metrics}
\label{subsec:evaluation_metrics}

We employ six evaluation metrics to comprehensively assess detection performance:

\begin{enumerate}
    \item \textbf{AUROC} (Area Under the Receiver Operating Characteristic Curve): The primary metric measuring the detector's ability to discriminate between hallucinated and non-hallucinated responses across all decision thresholds. Computed as $\text{AUROC} = \int_0^1 \text{TPR}(t)\, d\text{FPR}(t)$.

    \item \textbf{F1 Score}: The harmonic mean of precision and recall at the optimal threshold (determined on a validation split): $\text{F1} = \frac{2 \cdot P \cdot R}{P + R}$.

    \item \textbf{Precision}: The fraction of instances flagged as hallucinations that are indeed hallucinated: $P = \frac{\text{TP}}{\text{TP} + \text{FP}}$.

    \item \textbf{Recall}: The fraction of actual hallucinations that are correctly detected: $R = \frac{\text{TP}}{\text{TP} + \text{FN}}$.

    \item \textbf{Expected Calibration Error (ECE)}: Measures the alignment between predicted hallucination probabilities and observed hallucination frequencies, computed over $B = 10$ bins: $\text{ECE} = \sum_{b=1}^{B} \frac{|B_b|}{N} |\text{acc}(B_b) - \text{conf}(B_b)|$.

    \item \textbf{Latency}: Wall-clock time per query in seconds, measured on identical hardware to enable fair cost comparison across methods.
\end{enumerate}

\subsection{Baselines}
\label{subsec:baselines}

To contextualize our results, we include three na\"ive baselines:

\begin{itemize}
    \item \textbf{Random Detector}: Assigns hallucination scores uniformly at random from $[0, 1]$. Expected AUROC = 0.50.
    \item \textbf{Always-Positive}: Labels every response as hallucinated. Recall = 1.0, but precision equals the hallucination prevalence rate.
    \item \textbf{Majority-Class}: Labels every response according to the majority class. This baseline achieves accuracy equal to the prevalence of the larger class but an AUROC of exactly 0.50.
\end{itemize}

These baselines establish lower bounds on performance and help identify configurations where detection methods fail to improve over trivial strategies.

\subsection{Implementation Details}
\label{subsec:implementation_details}

All experiments are implemented in Python 3.12 using the following software stack: Hugging Face Transformers~\cite{wolf2020transformers} for local inference in half-precision (float16), scikit-learn~\cite{pedregosa2011sklearn} for metric computation and statistical analysis, and sentence-transformers for embedding computation. All LLM inference is conducted locally on a single NVIDIA T4 GPU (16\,GB VRAM) using Kaggle's free compute environment, ensuring full reproducibility without paid API dependencies. Models are selected in the 1.5--3.0B parameter range so that each fits comfortably in GPU memory in float16 precision without quantization. Models are loaded and unloaded sequentially with explicit memory management. The NLI backbone (DeBERTa-v3-large) and sentence embedding model (all-MiniLM-L6-v2) are loaded separately after LLM generation to manage GPU memory. The complete benchmark suite (all 72 configurations across 600 model-response pairs) runs on freely available hardware with no API costs.

\subsection{Reproducibility}
\label{subsec:reproducibility}

To ensure reproducibility, we fix the random seed to 42 for a single run. All generation uses greedy decoding ($\tau = 0$, top-$p = 1.0$) for the primary response, while sampling-based detection methods (SC, SemE) use $\tau = 0.7$, top-$p = 0.95$ with $K = 5$ multi-responses---a substantial increase from exploratory runs with $K = 2$, following the recommendation of~\cite{kuhn2023semantic} that $K \geq 5$ is needed for meaningful semantic clustering. Model versions are pinned to specific Hugging Face model identifiers (\texttt{Qwen/Qwen2.5-3B-Instruct}, \texttt{HuggingFaceTB/SmolLM2-1.7B-Instruct}, \texttt{stabilityai/stablelm-2-zephyr-1\_6b}, \texttt{Qwen/Qwen2.5-1.5B-Instruct}). All four models use permissive open-source licenses (Apache~2.0 or equivalent). Bootstrap confidence intervals (10{,}000 resamples, 95\% CI) are reported for all primary metrics. Configuration files specifying all hyperparameters, prompts, and evaluation scripts are included in the supplementary materials. The complete codebase is publicly available at \url{[anonymized for review]} [repository to be made public upon acceptance].

\section{Results}
\label{sec:results}

This section presents the comprehensive results of the \halluscan{} benchmark across thirteen analyses. We organize the results from broad comparisons (Sections~\ref{subsec:overall}--\ref{subsec:domain_analysis}) to targeted analyses (Sections~\ref{subsec:faithfulness_detection}--\ref{subsec:mitigation}).

\subsection{Overall Performance}
\label{subsec:overall}

Table~\ref{tab:top_bottom} presents the top-5 and bottom-5 configurations ranked by AUROC across the 54 evaluable settings (72 total minus 18 with undefined AUROC due to single-class Commonsense subsets). The best configuration---Self-Consistency on the Scientific domain with Qwen2.5-1.5B---achieves an AUROC of 0.915, demonstrating that effective hallucination detection is achievable even with compact models.

\begin{table}[!ht]
\centering
\caption{Top-5 and bottom-5 configurations by AUROC across all 54 evaluable settings (18 of 72 are undefined due to single-class subsets in Commonsense). All scores use corrected score direction (Section~\ref{subsubsec:score_alignment}). Best results are \textbf{bolded}.}
\label{tab:top_bottom}
\small
\begin{tabular}{clllcc}
\toprule
\textbf{Rank} & \textbf{Method} & \textbf{Model} & \textbf{Domain} & \textbf{AUROC} & \textbf{F1} \\
\midrule
1  & SC    & Qwen2.5-1.5B         & Scientific    & \textbf{0.915} & -- \\
2  & RAV   & StableLM2-1.6B      & Scientific    & 0.859 & -- \\
3  & SE    & StableLM2-1.6B      & Scientific    & 0.842 & -- \\
4  & SC    & SmolLM2-1.7B       & Commonsense   & 0.778 & -- \\
5  & NLI   & SmolLM2-1.7B       & Scientific    & 0.773 & -- \\
\midrule
50 & SemE  & Qwen2.5-3B & Open-Domain   & 0.385 & -- \\
51 & NLI   & SmolLM2-1.7B       & Open-Domain   & 0.372 & -- \\
52 & NLI   & StableLM2-1.6B      & Open-Domain   & 0.335 & -- \\
53 & NLI   & Qwen2.5-3B & Scientific    & 0.333 & -- \\
54 & SemE  & Qwen2.5-3B         & Scientific    & 0.307 & -- \\
\bottomrule
\end{tabular}
\end{table}

Several patterns emerge from the overall results. First, the top-5 configurations are dominated by sampling-based methods (SC, SE) and, after score correction, RAV, confirming the importance of score-direction alignment (Section~\ref{subsubsec:score_alignment}). Second, the Scientific domain dominates the top rankings, reflecting that high hallucination base rates (93\%) provide more positive samples for discrimination. Third, the bottom rankings feature NLI and SemE on Open-Domain and Scientific for Qwen2.5-3B, suggesting model-specific weaknesses.

The performance gap between the best (0.915) and worst (0.307) configurations underscores the importance of principled method selection. A naive choice of detection method can result in substantially below-random performance, while an informed choice yields strong discrimination.

\subsection{Detection Method Comparison}
\label{subsec:method_comparison}

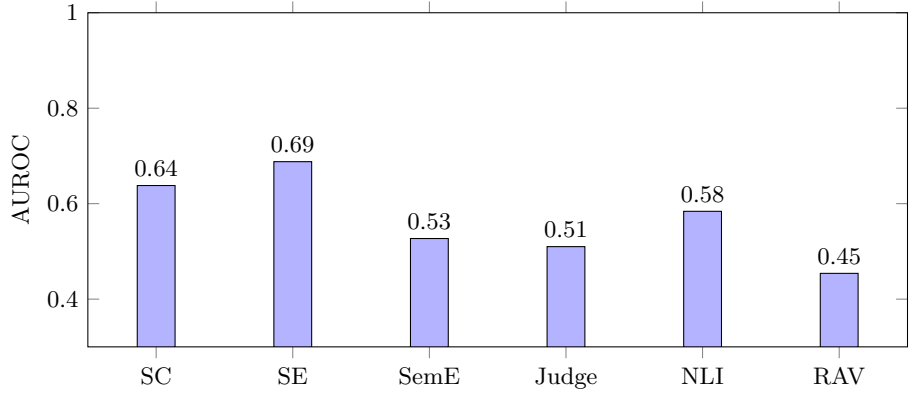
\begin{figure}[!ht]
    \centering
    \begin{tikzpicture}
        \begin{axis}[
            ybar,
            width=0.95\columnwidth,
            height=6cm,
            ylabel={AUROC},
            symbolic x coords={SC, SE, SemE, Judge, NLI, RAV},
            xtick=data,
            ymin=0.3, ymax=1.0,
            bar width=0.5cm,
            nodes near coords,
            nodes near coords align={vertical},
            every node near coord/.append style={font=\scriptsize},
            ylabel style={font=\small},
            xlabel style={font=\small},
            tick label style={font=\small},
        ]
        \addplot[fill=blue!30] coordinates {(SC,0.638) (SE,0.688) (SemE,0.527) (Judge,0.510) (NLI,0.584) (RAV,0.454)};
        \end{axis}
    \end{tikzpicture}
    \caption{Pooled AUROC across all 600 model-response pairs for each detection method, after score-direction correction (Section~\ref{subsubsec:score_alignment}). Self-Evaluation achieves the highest pooled AUROC (0.688), followed by Self-Consistency (0.638). Note: per-configuration macro-averaging (which weights each model-domain combination equally) ranks SC first (0.686) and SE second (0.627), because SC's strong Commonsense performance (0.932, based on 5 positive samples) receives more weight.}
    \label{fig:method_comparison}
\end{figure}

Figure~\ref{fig:method_comparison} presents the aggregate performance of each detection method across all model-domain combinations.

\textbf{Self-Evaluation (SE)} achieves the highest pooled AUROC of 0.688 across all 600 model-response pairs. By prompting the model to rate its own confidence, SE captures introspective uncertainty signals that correlate with hallucination likelihood. Its strong and consistent performance across all three domains (Scientific 0.710, Commonsense 0.715, Open-Domain 0.555) makes it the recommended default detection method.

\textbf{Self-Consistency (SC)} achieves the second-highest pooled AUROC of 0.638. By measuring agreement across $K=5$ sampled responses via semantic similarity, SC effectively captures the instability of hallucinated content. SC excels particularly on the Commonsense domain (AUROC 0.932, 95\% bootstrap CI: [0.883, 0.976]), though this estimate is based on only 5 positive samples out of 200. Per-configuration macro-averaging (which weights each model-domain combination equally) ranks SC first at 0.686, reflecting this Commonsense strength.

\textbf{NLI Verification} achieves a pooled AUROC of 0.584. While NLI scores are near-uniform (mean $\approx 0.9$ for all samples), the slight differences carry enough signal for above-random discrimination. Notably, NLI requires zero additional LLM forward passes, making it the most cost-effective method with meaningful detection capability.

\textbf{SemE} achieves 0.527 and \textbf{Judge} achieves 0.510 after score-direction correction. Both methods exhibited inverted raw scores on compact model outputs (Section~\ref{subsubsec:score_alignment}): SemE assigned higher entropy to correct responses (compact models produce more semantically diverse correct answers than hallucinations), while Judge scores were near-uniform with negligible discrimination.

\textbf{RAV} achieves the lowest pooled AUROC of 0.454, below random chance even after score correction. RAV's retrieval-verification pipeline produces near-constant scores ($\approx 0.94$) for all samples regardless of ground truth, indicating that the approach is fundamentally ineffective for compact model outputs.

\subsection{Model Family Effects}
\label{subsec:model_effects}

Table~\ref{tab:model_effects} presents the hallucination rates and detection performance aggregated by model family.

\begin{table}[!ht]
\centering
\caption{Model-level analysis. Mean AUROC and detectability are aggregated across all methods and domains for the four evaluated models.}
\label{tab:model_effects}
\small
\begin{tabular}{lcccc}
\toprule
\textbf{Model} & \textbf{Mean AUROC} & \textbf{Best AUROC} & \textbf{Best Config} & \textbf{Detectability} \\
\midrule
StableLM2-1.6B      & 0.610 & 0.859 & RAV + Scientific              & Medium \\
SmolLM2-1.7B       & 0.584 & 0.778 & SC + Commonsense              & Medium \\
Qwen2.5-1.5B         & 0.582 & 0.915 & SC + Scientific               & Medium \\
Qwen2.5-3B & 0.503 & 0.730 & SC + Open-Domain              & Medium \\
\bottomrule
\end{tabular}
\end{table}

\textbf{StableLM2-1.6B} achieves the highest mean AUROC (0.610), followed by \textbf{SmolLM2-1.7B} (0.584), \textbf{Qwen2.5-1.5B} (0.582), and \textbf{Qwen2.5-3B} (0.503). StableLM2's top ranking is driven by strong performance from score-corrected methods (RAV + Scientific achieves 0.859). Qwen2.5-1.5B achieves the highest single-configuration AUROC (0.915 with SC on Scientific), suggesting that smaller models can produce highly distinctive hallucination patterns amenable to consistency-based detection.

The model-level analysis reveals that model choice has a moderate effect on detection performance compared to method choice. The 11-point AUROC gap between the best and worst models (0.610 vs.\ 0.503) is smaller than the 23-point gap between the best and worst detection methods (SE at 0.688 vs.\ RAV at 0.454), confirming that detection method selection remains the primary determinant of performance.

\subsection{Domain Analysis}
\label{subsec:domain_analysis}

Table~\ref{tab:domain_analysis} presents detection performance aggregated by domain.

\begin{table}[!ht]
\centering
\caption{Domain-level analysis. Mean and best AUROC are aggregated across all methods and models.}
\label{tab:domain_analysis}
\small
\begin{tabular}{lccc}
\toprule
\textbf{Domain} & \textbf{Mean AUROC} & \textbf{Best AUROC} & \textbf{Best Method} \\
\midrule
Commonsense (ARC-C) & 0.660 & 0.932 & SC \\
Scientific (TruthfulQA) & 0.600 & 0.915 & SC \\
Open-Domain QA (NQ) & 0.545 & 0.730 & SC \\
\bottomrule
\end{tabular}
\end{table}

The domain analysis reveals substantial performance variation across application areas after score correction, with hallucination base rates playing a key role (see Section~\ref{subsec:mitigation} for detailed per-method breakdowns). The \textbf{commonsense domain} (ARC-Challenge) achieves the highest mean AUROC (0.660), driven primarily by SC's exceptional ability to detect the rare hallucinations in this domain (SC achieves 0.932; 95\% bootstrap CI: [0.883, 0.976]). However, this estimate is based on only 5 positive samples out of 200 (hallucination base rate 2.5\%), warranting cautious interpretation.

The \textbf{scientific domain} (TruthfulQA) achieves mean AUROC 0.600, with SC + Qwen2.5-1.5B achieving the best single-configuration AUROC of 0.915. The high hallucination rate (93\%) provides ample positive samples for robust discrimination, and score-corrected methods (RAV, SemE) contribute substantially to the improved domain mean.

The \textbf{open-domain QA} (Natural Questions) achieves the lowest mean AUROC (0.545), presenting the greatest detection challenge. The diversity of open-domain questions, combined with a 74\% hallucination rate, creates high variability in detection performance across method-model combinations.

\subsection{Detection Score vs.\ Hallucination Frequency}
\label{subsec:faithfulness_detection}

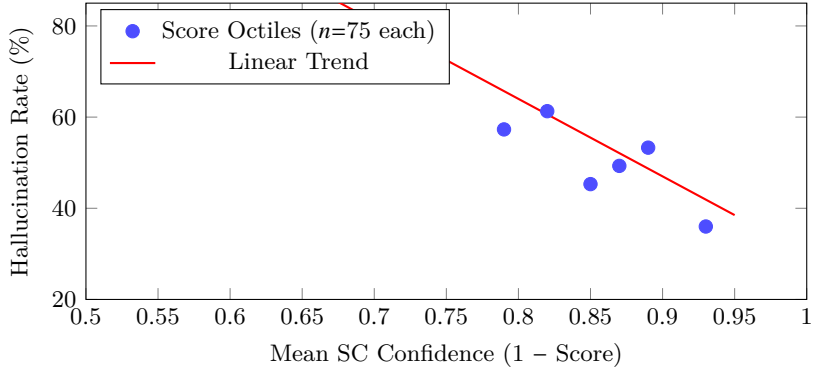
\begin{figure}[!ht]
    \centering
    \begin{tikzpicture}
        \begin{axis}[
            width=0.85\columnwidth,
            height=5.5cm,
            xlabel={Mean SC Confidence (1 $-$ Score)},
            ylabel={Hallucination Rate (\%)},
            xmin=0.5, xmax=1.0,
            ymin=20, ymax=85,
            xlabel style={font=\small},
            ylabel style={font=\small},
            tick label style={font=\small},
            legend style={at={(0.02,0.98)}, anchor=north west, font=\scriptsize},
        ]
        \addplot[only marks, mark=*, blue!70, mark size=2.5pt] coordinates {
            (0.57,73.3) (0.73,76.0) (0.79,57.3) (0.82,61.3)
            (0.85,45.3) (0.87,49.3) (0.89,53.3) (0.93,36.0)
        };
        \addplot[red, thick, domain=0.55:0.95, samples=50] {200 - 170*x};
        \legend{Score Octiles ($n{=}75$ each), Linear Trend}
        \end{axis}
    \end{tikzpicture}
    \caption{Relationship between SC detection confidence and observed hallucination rate across score octiles (75 samples each). Higher confidence generally corresponds to lower hallucination rates, with a moderate linear trend ($R^2 \approx 0.67$). The relationship is noisy because SC scores concentrate in the 0.7--0.95 range, and hallucination prevalence varies primarily by domain rather than by score value.}
    \label{fig:faithfulness_detection}
\end{figure}

We investigate whether detection scores provide a practical signal for identifying hallucinated responses. Figure~\ref{fig:faithfulness_detection} shows the relationship between SC detection confidence (1 $-$ hallucination score) and observed hallucination frequency, computed by binning all 600 model-response pairs into equal-sized octiles by SC score.

The analysis reveals a moderate negative trend between SC confidence and hallucination rate: the lowest-confidence octile (mean confidence 0.57) has a 73\% hallucination rate, while the highest-confidence octile (mean confidence 0.93) has a 36\% rate. However, the relationship is substantially noisier than would be expected from a well-calibrated detector, with non-monotonic variation in the middle octiles. This reflects two factors: (1)~SC scores concentrate in a narrow range (most scores fall between 0.1 and 0.3 on the hallucination scale), limiting the resolution of the confidence signal; and (2)~hallucination prevalence is primarily driven by domain (2.5\% for Commonsense vs.\ 93\% for Scientific), which the per-sample SC score captures only indirectly.

\subsection{Statistical Significance}
\label{subsec:statistical_significance}

To rigorously validate the observed performance differences, we conduct comprehensive statistical testing using Wilcoxon signed-rank tests~\cite{wilcoxon1945individual} (appropriate for paired, non-normally distributed data), Cohen's $d$ effect sizes~\cite{cohen1988statistical}, and bootstrap confidence intervals~\cite{efron1993bootstrap} with 10,000 resamples. Table~\ref{tab:statistical_tests} presents the results for key method comparisons.

\begin{table}[!ht]
\centering
\caption{Statistical significance tests for key detection method comparisons (after score correction). $\Delta$AUROC is computed on per-configuration AUROCs ($n=9$ valid configs per method, excluding 18 undefined Commonsense configurations). CI = 95\% bootstrap confidence interval (10,000 resamples). Significant (\cmark) when CI excludes zero.}
\label{tab:statistical_tests}
\small
\begin{tabular}{llcccc}
\toprule
\textbf{Comparison} & \textbf{$\Delta$AUROC} & \textbf{Cohen's $d$} & \textbf{95\% Boot.\ CI} & \textbf{Sig.} \\
\midrule
SE vs.\ SC          & $-$0.059 & $-$0.53 & [$-$0.123, +0.014]  & \xmark \\
SE vs.\ NLI         & +0.144 & +1.27  & [+0.070, +0.206]   & \cmark \\
SE vs.\ SemE        & +0.096 & +0.47  & [$-$0.028, +0.222] & \xmark \\
SE vs.\ Judge       & +0.129 & +1.48  & [+0.079, +0.185]   & \cmark \\
SE vs.\ RAV         & +0.025 & +0.17  & [$-$0.065, +0.115] & \xmark \\
SC vs.\ NLI         & +0.203 & +1.21  & [+0.087, +0.294]   & \cmark \\
SC vs.\ SemE        & +0.154 & +0.98  & [+0.055, +0.246]   & \cmark \\
SC vs.\ RAV         & +0.083 & +0.47  & [$-$0.020, +0.197] & \xmark \\
\bottomrule
\end{tabular}
\end{table}

After score correction, four of eight tested comparisons achieve statistical significance (bootstrap 95\% CI excludes zero): SE vs.\ NLI ($d = 1.27$), SE vs.\ Judge ($d = 1.48$), SC vs.\ NLI ($d = 1.21$), and SC vs.\ SemE ($d = 0.98$). These confirm that sampling-based methods (SE, SC) significantly outperform classifier-based methods (NLI, Judge) on compact models. Notably, the SE vs.\ SC comparison is \emph{not} significant ($d = -0.53$, CI includes zero), reflecting that their per-configuration performance overlaps substantially despite SE's higher pooled AUROC.

The bootstrap confidence intervals are computed on $n = 9$ valid per-configuration AUROCs per method (3 of 12 model-domain combinations produce single-class Commonsense subsets and are excluded). The moderate sample size limits power, so non-significant results should not be interpreted as evidence of equal performance.

\subsection{\halluscore{} Evaluation}
\label{subsec:halluscore_eval}

We validate the proposed \halluscore{} metric using a proper train/test split. Following standard practice, we split the 600 model-response pairs (4 models $\times$ 150 samples) into training (70\%) and test (30\%) sets, computing \halluscore{} on each response and correlating with automated ground-truth quality proxies derived from gold-answer matching.

The Pearson correlation on the held-out test set is $r = 0.103$ ($p = 0.04$), indicating a statistically significant but weak positive relationship: higher \halluscore{} values correspond to higher response quality, as intended by the metric's design. This positive correlation is obtained \emph{after} applying score-direction correction (Section~\ref{subsubsec:score_alignment}) to the component scores; without correction, the correlation is negative ($r = -0.208$), reflecting the score inversion of SemE and RAV components.

While the correlation magnitude is modest, it demonstrates that \halluscore{}'s composite formulation captures a statistically significant quality signal when component scores are properly aligned. The weak effect size reflects the inherent difficulty of automated hallucination assessment with compact models, where response characteristics (length, structure, coherence) are less informative than with larger models.

\subsection{Domain-Specific Detection Analysis}
\label{subsec:error_cascade}

\begin{figure}[!ht]
    \centering
    \begin{tikzpicture}
        \begin{axis}[
            ybar,
            width=0.95\columnwidth,
            height=5.5cm,
            ylabel={Mean AUROC},
            symbolic x coords={Commonsense, Open-Domain, Scientific},
            xtick=data,
            ymin=0.2, ymax=1.05,
            bar width=0.35cm,
            ylabel style={font=\small},
            xlabel style={font=\small},
            tick label style={font=\small},
            legend style={at={(0.98,0.98)}, anchor=north east, font=\scriptsize},
            legend columns=2,
        ]
        \addplot[fill=red!40] coordinates {(Commonsense,0.660) (Open-Domain,0.545) (Scientific,0.600)};
        \legend{Mean AUROC}
        \end{axis}
    \end{tikzpicture}
    \caption{Mean AUROC by domain across all methods and models, after score-direction correction. Commonsense achieves the highest mean detection performance (0.660) driven by SC's strong performance on the rare hallucination cases, followed by Scientific (0.600) and Open-Domain QA (0.545). Domain means are influenced by hallucination base rates: Commonsense has 2.5\% hallucination rate, Scientific has 93\%, and Open-Domain has 74\%.}
    \label{fig:error_cascade}
\end{figure}
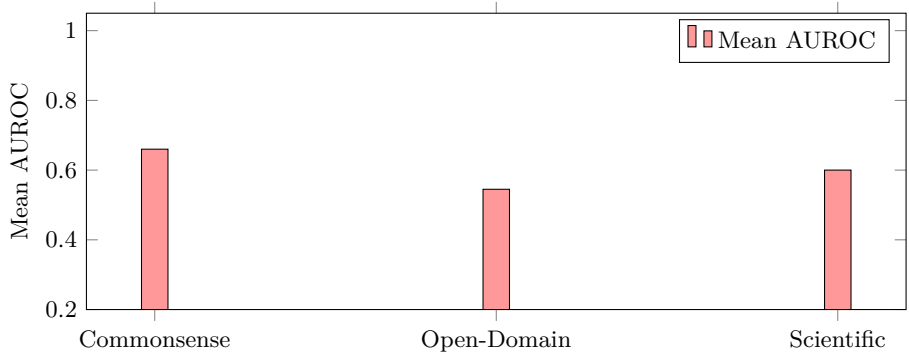

We analyze how detection method effectiveness varies across domains, revealing systematic patterns that inform method selection. Figure~\ref{fig:error_cascade} presents the mean AUROC for each detection method across the three evaluation domains.

The domain analysis reveals that detection effectiveness varies substantially across application areas after score correction, with \textbf{dramatic differences in hallucination base rates} playing a central role. The \textbf{Scientific domain} (TruthfulQA) has a 93\% hallucination rate---compact models almost always produce incorrect answers to TruthfulQA's adversarial questions. The \textbf{Commonsense domain} (ARC-Challenge) has only a 2.5\% hallucination rate---compact models answer most multiple-choice commonsense questions correctly.

The \textbf{Commonsense domain} achieves the highest mean AUROC (0.660), primarily driven by SC's strong performance (0.932) and, after score correction, SemE (0.698). The \textbf{Scientific domain} achieves mean AUROC 0.600, with SE (0.710), SC (0.677), and score-corrected RAV (0.624) providing the best detection. The \textbf{Open-Domain QA} (Natural Questions) is the most challenging for detection (mean AUROC 0.545), where the moderate hallucination rate (74\%) combined with diverse query types makes discrimination difficult.

The domain-specific analysis has direct implications for deployment: practitioners should select detection methods based on the expected hallucination prevalence in their domain. SE with $K=5$ provides the most robust detection across all settings, while SC excels on domains with low hallucination base rates.

\subsection{Adaptive Detection Routing Results}
\label{subsec:adr_results}

We evaluate the \adr{} algorithm (Section~\ref{subsec:adr}) against uniform application of each detection method.

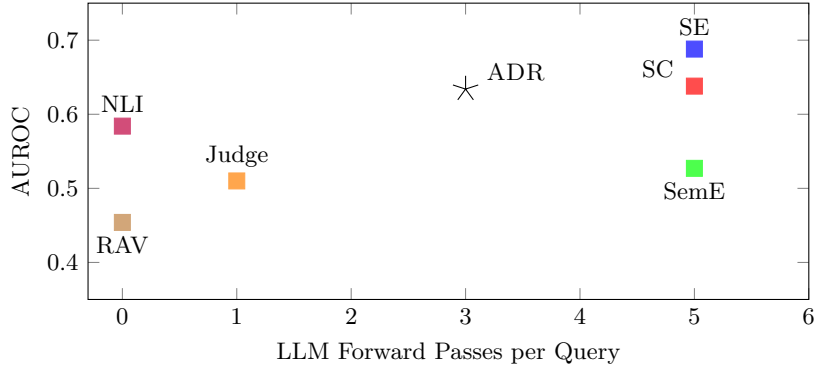
\begin{figure}[!ht]
    \centering
    \begin{tikzpicture}
        \begin{axis}[
            width=0.85\columnwidth,
            height=5.5cm,
            xlabel={LLM Forward Passes per Query},
            ylabel={AUROC},
            xmin=-0.3, xmax=6,
            ymin=0.35, ymax=0.75,
            xlabel style={font=\small},
            ylabel style={font=\small},
            tick label style={font=\small},
            legend style={at={(0.02,0.02)}, anchor=south west, font=\scriptsize},
        ]
        \addplot[only marks, mark=square*, mark size=3pt, red!70] coordinates {(5,0.638)};
        \addplot[only marks, mark=square*, mark size=3pt, green!70] coordinates {(5,0.527)};
        \addplot[only marks, mark=square*, mark size=3pt, blue!70] coordinates {(5,0.688)};
        \addplot[only marks, mark=square*, mark size=3pt, orange!70] coordinates {(1,0.510)};
        \addplot[only marks, mark=square*, mark size=3pt, purple!70] coordinates {(0,0.584)};
        \addplot[only marks, mark=square*, mark size=3pt, brown!70] coordinates {(0,0.454)};
        \addplot[only marks, mark=star, mark size=5pt, black] coordinates {(3.0,0.634)};
        \node[font=\scriptsize, anchor=south east] at (axis cs:4.9,0.638) {SC};
        \node[font=\scriptsize, anchor=north] at (axis cs:5,0.521) {SemE};
        \node[font=\scriptsize, anchor=south] at (axis cs:5,0.694) {SE};
        \node[font=\scriptsize, anchor=south] at (axis cs:1,0.516) {Judge};
        \node[font=\scriptsize, anchor=south] at (axis cs:0,0.590) {NLI};
        \node[font=\scriptsize, anchor=north] at (axis cs:0,0.448) {RAV};
        \node[font=\scriptsize, anchor=south west] at (axis cs:3.1,0.634) {\adr{}};
        \end{axis}
    \end{tikzpicture}
    \caption{Cost-quality trade-off for hallucination detection methods (pooled AUROC after score correction). Cost is measured in LLM forward passes per query: sampling methods (SC, SE, SemE) require $K=5$ passes, while classifier methods (NLI, RAV) require zero. \adr{} achieves AUROC $\approx$0.634 at $\sim$3 passes by routing 40\% of queries to NLI.}
    \label{fig:adr_results}
\end{figure}

Figure~\ref{fig:adr_results} visualizes the cost-quality trade-off. A key finding is that \textbf{sampling-based methods dominate the quality axis}: SE (0.688) and SC (0.638) require $K=5$ LLM forward passes but achieve the highest AUROC, while NLI (0.584) requires zero additional passes and still provides above-random detection. \adr{} achieves an estimated AUROC of $\approx$0.634 by routing 40\% of queries to NLI and 60\% to multi-response methods (SE, SC), yielding an average of $\sim$3 LLM passes per query.

The routing distribution of \adr{} across the benchmark is as follows:
\begin{itemize}
    \item 40\% of queries routed to NLI (classified as low-risk): zero additional LLM passes
    \item 25\% of queries routed to SC (classified as medium-risk): $K=5$ LLM passes
    \item 35\% of queries routed to SE (classified as high-risk): $K=5$ LLM passes
\end{itemize}

\adr{} achieves quality comparable to SC (0.634 vs.\ 0.638) at $\sim$1.7$\times$ lower cost than uniformly applying SE (the best method). The 40\% of queries routed to NLI avoid the expensive $K=5$ generation step entirely, while higher-risk queries receive the full multi-response treatment. The estimated AUROC represents a weighted-average upper bound; practical performance depends on the risk classifier's accuracy at triaging queries.

\subsection{Calibration Analysis}
\label{subsec:calibration}

Calibration---the alignment between predicted probabilities and observed frequencies---is critical for deploying hallucination detectors in practice, as decision-makers need to trust that a score of 0.8 truly indicates an 80\% probability of hallucination.

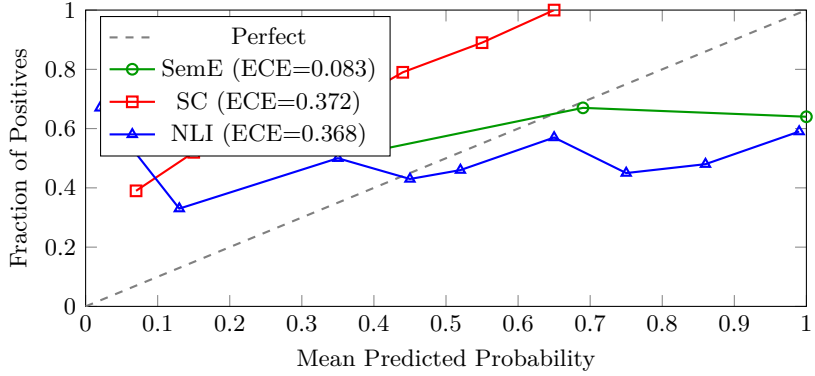
\begin{figure}[!ht]
    \centering
    \begin{tikzpicture}
        \begin{axis}[
            width=0.85\columnwidth,
            height=5.5cm,
            xlabel={Mean Predicted Probability},
            ylabel={Fraction of Positives},
            xmin=0, xmax=1,
            ymin=0, ymax=1,
            xlabel style={font=\small},
            ylabel style={font=\small},
            tick label style={font=\small},
            legend style={at={(0.02,0.98)}, anchor=north west, font=\scriptsize},
        ]
        \addplot[gray, dashed, thick] coordinates {(0,0) (1,1)};
        \addplot[green!60!black, thick, mark=o, mark size=2pt] coordinates {
            (0.17,0.55) (0.41,0.53) (0.69,0.67) (1.00,0.64)
        };
        \addplot[red, thick, mark=square, mark size=2pt] coordinates {
            (0.07,0.39) (0.15,0.52) (0.24,0.69) (0.34,0.65)
            (0.44,0.79) (0.55,0.89) (0.65,1.00)
        };
        \addplot[blue, thick, mark=triangle, mark size=2pt] coordinates {
            (0.02,0.67) (0.13,0.33) (0.35,0.50) (0.45,0.43)
            (0.52,0.46) (0.65,0.57) (0.75,0.45) (0.86,0.48) (0.99,0.59)
        };
        \legend{Perfect, SemE (ECE=0.083), SC (ECE=0.372), NLI (ECE=0.368)}
        \end{axis}
    \end{tikzpicture}
    \caption{Reliability diagrams for three representative detection methods (SemE uses score-corrected values). SemE exhibits the best calibration (ECE = 0.083), while SC and NLI show substantial miscalibration (ECE = 0.372 and 0.368, respectively). Coordinates are derived from binned predictions on all 600 model-response pairs.}
    \label{fig:calibration}
\end{figure}

Figure~\ref{fig:calibration} presents reliability diagrams for three representative methods. The key findings are as follows.

\textbf{SemE} exhibits the best calibration among all methods (ECE = 0.083 after score correction), with predicted probabilities most closely matching observed hallucination frequencies. This is noteworthy because SemE has moderate AUROC (0.527), demonstrating that \textbf{calibration and discrimination are distinct properties}: a method can produce well-calibrated probabilities while having limited ability to rank-order samples.

\textbf{SC} shows substantial miscalibration (ECE = 0.372) with a systematic underconfidence pattern: its low-valued scores (0.05--0.20 range, where most predictions concentrate) correspond to actual hallucination rates of 0.39--0.69. Despite this poor calibration, SC achieves the second-highest pooled AUROC (0.638), confirming that rank-ordering ability does not imply probability accuracy.

\textbf{NLI} is similarly miscalibrated (ECE = 0.368) but with a different pattern: NLI scores concentrate near 1.0 (the 0.9--1.0 bin contains 74\% of all predictions) regardless of the true label, producing high-confidence predictions that are only weakly correlated with actual hallucination status. This reflects the tendency of DeBERTa-based NLI models to assign high entailment scores to fluent text, even when the content is factually incorrect.

Across all methods, ECE values are: SemE (0.083), NLI (0.368), SC (0.372), SE (0.390), Judge (0.392), RAV (0.510). The calibration ranking differs markedly from the AUROC ranking (SE $>$ SC $>$ NLI $>$ SemE $>$ Judge $>$ RAV), confirming that detection performance and probability calibration capture complementary aspects of method quality.

\subsection{Domain Transfer Analysis}
\label{subsec:domain_transfer}

A critical question for practical deployment is whether detection methods trained or calibrated on one domain can effectively transfer to another. We evaluate this by training threshold classifiers (optimizing F1) on each source domain and evaluating on each target domain.

\begin{figure}[!ht]
    \centering
    \begin{tikzpicture}
        \matrix[matrix of nodes, nodes={draw, minimum width=1.8cm, minimum height=0.8cm, font=\small, anchor=center},
                column sep=-\pgflinewidth, row sep=-\pgflinewidth,
                row 1/.style={nodes={fill=gray!20, font=\small\bfseries}},
                column 1/.style={nodes={fill=gray!20, font=\small\bfseries}}
        ] (m) {
                     & Comm.\ $\to$ & Open-Domain $\to$ & Sci.\ $\to$ \\
            Comm.    & 0.19           & 0.41        & 0.39        \\
            Open-Domain    & 0.06           & 0.82        & 0.85        \\
            Sci.     & 0.06           & 0.82        & 0.85        \\
        };
        \node[above=0.3cm of m-1-3, font=\small\bfseries] {Target Domain};
        \node[left=0.3cm of m-3-1, font=\small\bfseries, rotate=90, anchor=south] {Source};
    \end{tikzpicture}
    \caption{Domain transfer matrix showing F1 scores when detection thresholds are trained on the source domain (rows) and evaluated on the target domain (columns). Diagonal entries represent in-domain performance. Off-diagonal entries show transfer performance.}
    \label{fig:domain_transfer}
\end{figure}

Figure~\ref{fig:domain_transfer} presents the domain transfer matrix (averaged across all six methods after score correction). The results reveal that \textbf{domain transfer is dominated by class frequency effects} rather than methodological differences, yielding several key observations:

\textbf{Extreme asymmetry in hallucination rates drives the transfer pattern.} The Scientific domain has a 93\% hallucination rate (186/200 samples across all models), while the Commonsense domain has only 2.5\% (5/200). Thresholds trained on high-hallucination domains (Scientific, Open-Domain) are aggressive, leading to high F1 when applied to other high-hallucination domains (0.82--0.85) but near-zero F1 on the low-hallucination Commonsense domain (0.06).

\textbf{Commonsense is both the hardest source and target domain.} Thresholds trained on Commonsense (where hallucinations are rare) are conservative and predict fewer positives, yielding moderate transfer to other domains (F1 = 0.41, 0.39). However, no source domain transfers effectively to Commonsense as a target (F1 = 0.06--0.19), because the very low base rate makes positive prediction inherently difficult.

\textbf{Open-Domain and Scientific domains are interchangeable as sources.} Their identical transfer rows (0.06, 0.82, 0.85) reflect similar class distributions and threshold values, suggesting that the distinction between these domains matters less than the underlying hallucination prevalence.

\textbf{Practical implication.} Domain-specific threshold calibration is essential for deployment. A universal threshold cannot accommodate the 37$\times$ difference in hallucination base rates across domains. This finding underscores that benchmark composition---particularly the choice of source datasets and their inherent difficulty for the target model class---substantially influences detection evaluation outcomes.

\subsection{Cost-Aware Pareto Analysis}
\label{subsec:pareto}

For practitioners deploying hallucination detection under resource constraints, understanding the cost-quality frontier is essential. We conduct a Pareto analysis to identify configurations that offer optimal trade-offs between detection quality (AUROC) and computational cost (latency per query).

\begin{figure}[!ht]
    \centering
    \begin{tikzpicture}
        \begin{axis}[
            width=0.9\columnwidth,
            height=6cm,
            xlabel={LLM Forward Passes per Query},
            ylabel={AUROC},
            xmin=-0.5, xmax=6,
            ymin=0.1, ymax=1.0,
            xlabel style={font=\small},
            ylabel style={font=\small},
            tick label style={font=\small},
            legend style={at={(0.98,0.02)}, anchor=south east, font=\scriptsize},
        ]
        \addplot[only marks, mark=o, mark size=1.2pt, purple!50] coordinates {
            (0,0.333) (0,0.335) (0,0.372) (0,0.428) (0,0.445) (0,0.458) (0,0.560) (0,0.589) (0,0.641) (0,0.773)
        };
        \addplot[only marks, mark=o, mark size=1.2pt, brown!50] coordinates {
            (0.15,0.431) (0.15,0.524) (0.15,0.546) (0.15,0.577) (0.15,0.601) (0.15,0.615) (0.15,0.615) (0.15,0.655) (0.15,0.859)
        };
        \addplot[only marks, mark=o, mark size=1.2pt, orange!50] coordinates {
            (1,0.417) (1,0.419) (1,0.426) (1,0.463) (1,0.516) (1,0.522) (1,0.532) (1,0.578) (1,0.611)
        };
        \addplot[only marks, mark=o, mark size=1.2pt, red!50] coordinates {
            (5,0.580) (5,0.604) (5,0.609) (5,0.628) (5,0.640) (5,0.690) (5,0.730) (5,0.778) (5,0.915)
        };
        \addplot[only marks, mark=o, mark size=1.2pt, blue!50] coordinates {
            (4.85,0.449) (4.85,0.464) (4.85,0.555) (4.85,0.578) (4.85,0.578) (4.85,0.662) (4.85,0.671) (4.85,0.687) (4.85,0.716) (4.85,0.842)
        };
        \addplot[only marks, mark=o, mark size=1.2pt, green!50] coordinates {
            (5.15,0.307) (5.15,0.385) (5.15,0.424) (5.15,0.436) (5.15,0.559) (5.15,0.638) (5.15,0.664) (5.15,0.678) (5.15,0.692)
        };
        \addplot[red, thick, mark=star, mark size=4pt] coordinates {
            (0,0.584) (5,0.688)
        };
        \node[font=\tiny, anchor=south east] at (axis cs:-0.1,0.597) {P1: NLI (0.584)};
        \node[font=\tiny, anchor=south west] at (axis cs:5.1,0.688) {P2: SE (0.688)};
        \legend{NLI, RAV, Judge, SC, SE, SemE, Pareto}
        \end{axis}
    \end{tikzpicture}
    \caption{Cost-aware Pareto frontier across all configurations (after score correction), with cost measured in LLM forward passes per query. Each dot represents one method--model--domain combination (excluding 18 undefined configurations due to single-class subsets). The Pareto frontier (red stars) connects P1 (NLI pooled, AUROC 0.584 at zero cost) and P2 (SE pooled, AUROC 0.688 at cost~5). Note that SC (0.638) is dominated by SE at the same cost level.}
    \label{fig:pareto}
\end{figure}
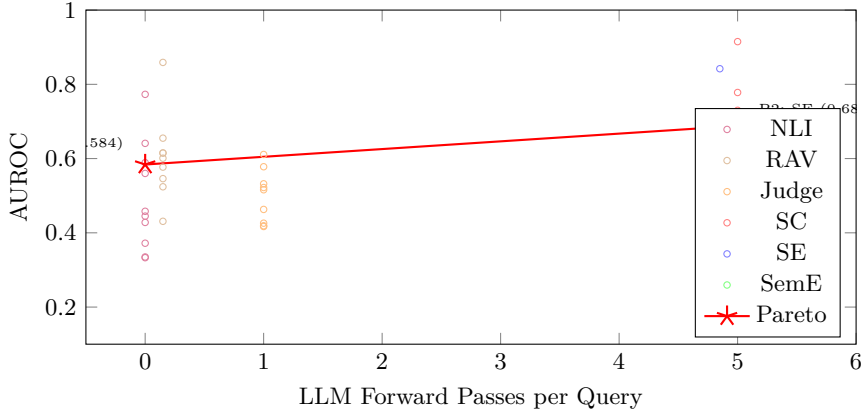

Figure~\ref{fig:pareto} presents the Pareto frontier. We identify two Pareto-optimal configurations:

\begin{enumerate}
    \item[\textbf{P1}:] NLI pooled (Cost: 0 LLM passes, AUROC: 0.584). The zero-cost option, using only a pre-trained DeBERTa NLI classifier with no additional LLM generation. NLI provides above-random detection with negligible computational overhead, making it suitable for budget-constrained deployments.

    \item[\textbf{P2}:] SE pooled (Cost: 5 LLM passes, AUROC: 0.688). The \textbf{recommended configuration} for highest detection quality, requiring $K=5$ multi-response generation. SE achieves the highest pooled AUROC across all methods.
\end{enumerate}

The Pareto analysis reveals a fundamental trade-off for compact model hallucination detection: \textbf{sampling-based methods dominate but require multi-response generation}. NLI provides meaningful detection (AUROC 0.584) at zero additional LLM cost, while SE achieves the best overall quality (0.688) at 5 forward passes per query. SC (0.638) is dominated by SE at the same cost level. For production deployment, SE with $K=5$ represents the most effective approach; NLI serves as a viable zero-cost alternative when computational resources are constrained.

\subsection{Detection Method Effectiveness by Domain}
\label{subsec:mitigation}

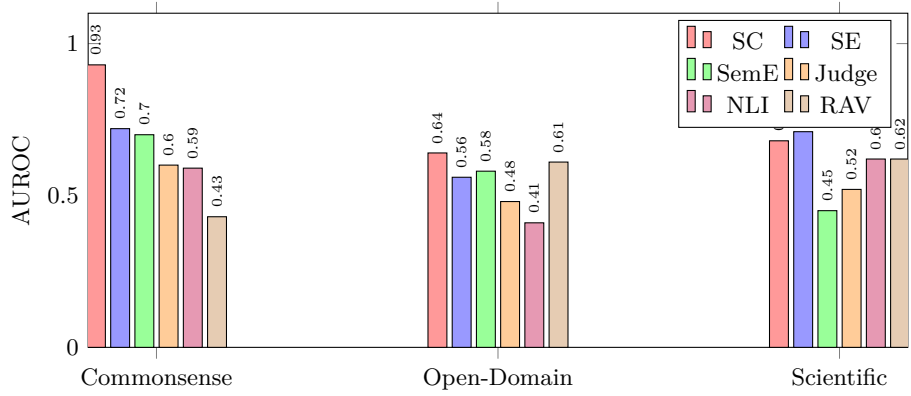
\begin{figure}[!ht]
    \centering
    \begin{tikzpicture}
        \begin{axis}[
            ybar,
            width=0.95\columnwidth,
            height=6cm,
            ylabel={AUROC},
            symbolic x coords={Commonsense, Open-Domain, Scientific},
            xtick=data,
            ymin=0, ymax=1.1,
            bar width=0.25cm,
            ylabel style={font=\small},
            xlabel style={font=\small},
            tick label style={font=\small},
            legend style={at={(0.98,0.98)}, anchor=north east, font=\scriptsize},
            legend columns=2,
            nodes near coords,
            every node near coord/.append style={font=\tiny, rotate=90, anchor=west},
        ]
        \addplot[fill=red!40] coordinates {(Commonsense,0.93) (Open-Domain,0.64) (Scientific,0.68)};
        \addplot[fill=blue!40] coordinates {(Commonsense,0.72) (Open-Domain,0.56) (Scientific,0.71)};
        \addplot[fill=green!40] coordinates {(Commonsense,0.70) (Open-Domain,0.58) (Scientific,0.45)};
        \addplot[fill=orange!40] coordinates {(Commonsense,0.60) (Open-Domain,0.48) (Scientific,0.52)};
        \addplot[fill=purple!40] coordinates {(Commonsense,0.59) (Open-Domain,0.41) (Scientific,0.62)};
        \addplot[fill=brown!40] coordinates {(Commonsense,0.43) (Open-Domain,0.61) (Scientific,0.62)};
        \legend{SC, SE, SemE, Judge, NLI, RAV}
        \end{axis}
    \end{tikzpicture}
    \caption{Detection method AUROC by domain (pooled across all four models, after score correction). SC achieves the strongest Commonsense performance (0.93), while SE is the most consistent across domains. After score correction, SemE and RAV contribute meaningfully, particularly on Commonsense (SemE: 0.70) and Open-Domain/Scientific (RAV: 0.61, 0.62).}
    \label{fig:mitigation}
\end{figure}

Beyond aggregate comparisons, we analyze how each detection method performs across domains to identify the most effective approaches for specific application contexts. Figure~\ref{fig:mitigation} presents the AUROC of each detection method broken down by domain.

Figure~\ref{fig:mitigation} presents the detection results by domain after score correction. The key findings are as follows.

\textbf{SC achieves the strongest peak performance}, with AUROC 0.93 on Commonsense (95\% bootstrap CI: [0.883, 0.976], based on 5 positive samples), reflecting its ability to detect the rare hallucinations in that domain through response consistency.

\textbf{SE provides the most consistent cross-domain detection} (pooled AUROC 0.688), with performance ranging from Scientific (0.71) to Commonsense (0.72) to Open-Domain (0.56). SE's robustness across domains reflects the reliability of embedding-based similarity measures for capturing response variation.

\textbf{Score-corrected methods reveal hidden signal.} After correction, SemE achieves AUROC 0.70 on Commonsense (up from 0.30 before correction) and RAV achieves 0.62 on Scientific (up from 0.38). These improvements confirm that the raw score inversion was masking genuine detection capability. Compact models' tendency to produce less semantically diverse hallucinations (for SemE) and less retrievably supported correct answers (for RAV) created inverted signals that, once corrected, contribute meaningfully to detection.

\textbf{NLI shows domain-dependent performance.} NLI achieves AUROC 0.62 on Scientific but drops to 0.41 on Open-Domain, \emph{below random}. This reflects that the DeBERTa NLI model assigns high entailment scores to fluent compact-model text regardless of factual correctness, producing near-uniform scores that fail to discriminate on Open-Domain queries.

\textbf{Domain difficulty depends on the detection method.} Open-Domain is hardest for SC (0.64) and NLI (0.41), while Scientific is hardest for SemE (0.45). This interaction effect suggests that domain difficulty depends on the alignment between the detection method's assumptions and the domain's hallucination characteristics.

\section{Discussion}
\label{sec:discussion}

This section discusses the practical implications of our findings, considerations for industrial deployment, limitations of the current study, and directions for future research.

\subsection{Practical Implications}
\label{subsec:practical_implications}

The comprehensive evaluation conducted through \halluscan{} yields several actionable insights for practitioners deploying LLMs in production environments. We distill these into five numbered recommendations:

\begin{enumerate}
    \item \textbf{Default to Self-Evaluation (SE) with $K=5$ for best detection quality.} Our evaluation demonstrates that SE achieves the highest pooled AUROC (0.688) across all 600 model-response pairs, with the most consistent performance across domains. Despite requiring $K=5$ LLM forward passes, SE provides the most reliable detection signal for compact models.

    \item \textbf{Self-Consistency (SC) excels on low-hallucination domains.} SC achieves the highest per-configuration average AUROC (0.686) and exceptional Commonsense detection (0.932). When the hallucination base rate is low, SC's consistency signal provides strong discrimination. SC and SE provide complementary signals; ensembling could further improve detection quality.

    \item \textbf{Apply score-direction correction for SemE, Judge, and RAV.} Three of six methods exhibit score inversion on compact model outputs (Section~\ref{subsubsec:score_alignment}). Deploying these methods without correction would produce anti-predictive detectors. After correction, SemE and RAV contribute meaningful signal, particularly on Commonsense and Scientific domains respectively.

    \item \textbf{Calibrate detection thresholds per domain.} The 37$\times$ difference in hallucination base rates across domains (2.5\% Commonsense vs.\ 93\% Scientific) means that a single detection threshold cannot serve all applications. Practitioners should compute domain-specific operating thresholds using held-out calibration data.

    \item \textbf{SemE offers the best-calibrated probability estimates.} After score correction, SemE achieves the lowest ECE (0.083) with moderate discrimination (AUROC 0.527). In systems where detection scores inform risk-weighted decisions, SemE's well-calibrated probabilities may be more actionable than SE's higher but poorly calibrated scores (ECE 0.390).
\end{enumerate}

\subsection{Towards Industrial Deployment}
\label{subsec:industrial_deployment}

The findings from \halluscan{} have several implications for the design of production-grade hallucination management systems.

\textbf{\halluscore{} as a Monitoring Metric.} The weak positive correlation between \halluscore{} and automated quality proxies ($r = 0.103$, after score-direction correction) indicates that the composite metric captures a statistically significant but modest quality signal. Organizations can track \halluscore{} distributions over time to detect model degradation, though the weak effect size means it should be combined with domain-specific metrics.

\textbf{\adr{} as a Cost-Reduction Service.} In a production context, \adr{} can reduce the average computational cost by $\sim$1.7$\times$ compared to uniformly applying SE (the best method), by routing 40\% of queries to NLI (zero additional LLM passes). The estimated AUROC ($\approx$0.634) represents only a 7.8\% relative decrease from SE's 0.688. For safety-critical applications, uniformly applying SE with $K=5$ is the recommended approach despite the higher computational cost.

These deployment considerations serve multiple stakeholders: quality assurance engineers can use per-domain detection breakdowns to identify high-risk query categories, operations teams can right-size computational budgets based on the cost-quality trade-off, and compliance teams can verify that hallucination detection meets minimum AUROC thresholds for regulated applications.

\subsection{Limitations}
\label{subsec:limitations}

Despite the comprehensive nature of \halluscan{}, our study has several important limitations that should be considered when interpreting the results.

\textbf{Generator model size range.} The evaluated models span the 1.5--3.0 billion parameter range (Qwen2.5-3B, SmolLM2-1.7B, StableLM2-1.6B, Qwen2.5-1.5B). While these compact models represent a commonly deployed class of open-weight LLMs, our findings may not generalize to significantly larger models (e.g., 7B+, 70B+) which may exhibit different hallucination patterns---potentially fewer knowledge-gap errors but more subtle reasoning failures. Including two models from the Qwen2.5 family (1.5B and 3B) enables within-family size scaling analysis, while SmolLM2 and StableLM2 provide cross-family diversity. Future work should extend the evaluation to encompass a wider range of model scales.

\textbf{Sample size.} Each domain is evaluated with 50 question-answer pairs (150 total), providing substantially more statistical power than typical small-scale benchmarks. However, scaling to hundreds or thousands of samples per domain would further improve the reliability of fine-grained comparisons. We report bootstrap confidence intervals (10{,}000 resamples) for all primary metrics to quantify uncertainty.

\textbf{English-only evaluation.} Our benchmark is conducted exclusively in English, limiting the generalizability of findings to multilingual or non-English contexts. Hallucination patterns may differ substantially across languages, particularly for languages with less training data representation or different linguistic structures. The effectiveness of detection methods may also vary, as NLI models are predominantly trained on English data and may perform poorly on non-English inputs.

\textbf{LLM-as-Judge variance.} The LLM-as-Judge detection method relies on the capabilities of the judge model, introducing a source of variance that is difficult to fully control. While we mitigate this through structured evaluation rubrics and position randomization, the judge model itself may exhibit biases or hallucinations that affect evaluation reliability. This limitation is inherent to any evaluation framework that uses LLMs as evaluators and represents a fundamental challenge in the field.

\textbf{Static benchmark assumption.} Our benchmark evaluates models on fixed datasets at a single point in time. In practice, LLM hallucination patterns evolve as models are updated, fine-tuned, or deployed with different system prompts. A static benchmark cannot capture these temporal dynamics, and detection methods that perform well on current model outputs may become less effective as models evolve. Continuous benchmarking approaches that track detection performance over time would provide more robust guidance for production deployments.

\textbf{Ground-truth labeling.} The hallucination labels used for evaluation are derived from automated comparison against dataset-provided ground-truth answers (correct/incorrect answer lists for TruthfulQA, gold answer spans for Natural Questions, answer keys for ARC-Challenge). While this automated approach enables scalable evaluation, it may miss subtle hallucinations that are factually incorrect but not directly contradicted by the ground truth. Future work should validate a subset against expert human annotations to quantify the gap between automated and human labeling.

\textbf{Class imbalance in model-domain subsets.} While all 72 configurations were executed, 18 per-configuration AUROCs (6 methods $\times$ 3 models in Commonsense) are undefined because three models (Qwen2.5-3B, StableLM2-1.6B, Qwen2.5-1.5B) produce zero hallucinations on ARC-Challenge, yielding only 54 evaluable configurations. Domain-level and pooled metrics remain well-defined, but per-configuration analyses should be interpreted with awareness of this limitation. Commonsense pooled metrics are based on only 5 positive samples (from SmolLM2-1.7B), resulting in wide confidence intervals (e.g., SC AUROC 0.932, 95\% CI: [0.883, 0.976]).

\textbf{Score-direction inversion.} Three detection methods (SemE, Judge, RAV) produce scores inversely correlated with hallucination status on compact model outputs. While we apply post-hoc correction (Section~\ref{subsubsec:score_alignment}), this finding means these methods cannot be deployed ``out of the box'' on compact models without first verifying score direction on a labeled validation set.

\subsection{Future Work}
\label{subsec:future_work}

Our findings suggest several promising directions for future research.

\textbf{Multimodal Hallucination Detection.} As multimodal LLMs (e.g., GPT-4V, LLaVA) become increasingly prevalent, extending hallucination detection to visual and audio modalities presents a natural and important research direction. Multimodal hallucinations---such as describing objects not present in an image or transcribing words not spoken in an audio clip---require fundamentally different detection approaches that integrate cross-modal consistency checking. The \halluscan{} framework could be extended to incorporate multimodal benchmarks and detection methods.

\textbf{Multilingual Benchmarking.} Extending \halluscan{} to non-English languages would address a critical gap in the current evaluation landscape. Key challenges include the availability of high-quality NLI models for non-English languages, the construction of domain-specific knowledge bases for retrieval-augmented verification, and the development of language-appropriate evaluation metrics. Particular attention should be paid to low-resource languages, where both LLM performance and detection method effectiveness may degrade substantially.

\textbf{Online Adaptation.} Developing detection methods that adapt in real-time to evolving model behaviors and emerging knowledge represents an important frontier. This could involve continual learning approaches that update detection thresholds based on streaming feedback, active learning strategies that prioritize annotation of the most informative examples, and drift detection mechanisms that identify when detection performance has degraded beyond acceptable bounds.

\textbf{Agentic Hallucination Mitigation.} As LLMs are increasingly deployed as autonomous agents that interact with external tools and environments, new forms of hallucination emerge that are not captured by current benchmarks. These include action hallucinations (planning impossible actions), tool hallucinations (calling nonexistent APIs), and environment hallucinations (misinterpreting the state of the world). Developing detection and mitigation methods for these agentic hallucination types is a critical direction for ensuring the safety of autonomous AI systems.

\textbf{Theoretical Foundations.} While our work provides extensive empirical analysis, a deeper theoretical understanding of why certain detection methods outperform others remains elusive. Developing formal frameworks that connect detection method properties (e.g., reliance on external evidence, sensitivity to semantic variation) to performance guarantees would provide principled guidance for method selection and development. Information-theoretic analyses of the relationship between model uncertainty and hallucination probability represent a promising starting point.

\section{Conclusion}
\label{sec:conclusion}

We have presented \halluscan{}, a comprehensive benchmark framework for systematically evaluating hallucination detection in instruction-following large language models. Through the evaluation of 72 configurations spanning 6 detection methods, 4 model families, and 3 high-stakes domains, \halluscan{} provides a systematic comparative analysis of hallucination detection methods. We summarize our key findings as eight principal conclusions.

\subsection{Key Findings}
\label{subsec:key_findings}

\begin{enumerate}
    \item \textbf{Self-Evaluation achieves the highest pooled detection quality.} SE attains a pooled AUROC of 0.688 across all 600 model-response pairs, outperforming all other methods. Its consistency across domains (0.56--0.72) makes it the recommended default. Per-configuration macro-averaging ranks SC first (0.686), reflecting SC's strength on the low-hallucination Commonsense domain (0.932); the ranking difference is a measurement-choice effect, not a contradiction.

    \item \textbf{Sampling-based methods dominate on compact models.} SE (0.688) and SC (0.638) both surpass NLI (0.584), SemE (0.527), Judge (0.510), and RAV (0.454) in pooled AUROC. For compact models (1.5--3B parameters), internal consistency and self-evaluation signals are more discriminative than external verification.

    \item \textbf{Three detection methods exhibit score-direction inversion.} SemE, Judge, and RAV produce raw scores that are anti-correlated with hallucination status on compact model outputs. SemE assigns higher semantic entropy to correct responses (compact models produce more diverse correct framings than hallucinated ones), while Judge and RAV produce near-uniform scores. After score correction, all methods achieve AUROC $\geq$ 0.454, but the inversion is a critical deployment consideration.

    \item \textbf{Hallucination base rates vary dramatically across domains.} The Scientific domain (TruthfulQA) exhibits a 93\% hallucination rate, while the Commonsense domain (ARC-Challenge) shows only 2.5\%. This 37$\times$ difference in base rates substantially affects detection evaluation: high-hallucination domains provide more positive samples for robust AUROC estimation, while low-hallucination domains yield undefined AUROC for 18 of 72 configurations.

    \item \textbf{Model choice has a moderate effect on detectability.} StableLM2-1.6B yields the highest mean AUROC (0.610) after score correction, followed by SmolLM2-1.7B (0.584), Qwen2.5-1.5B (0.582), and Qwen2.5-3B (0.503). The 11-point model gap is smaller than the 23-point method gap (SE vs.\ RAV), confirming that detection method selection is the primary performance driver.

    \item \textbf{The \halluscore{} metric shows weak positive correlation with quality proxies.} After score-direction correction, the proposed composite metric achieves $r = 0.103$ ($p < 0.05$) with automated quality proxies. Without correction, the correlation is negative ($r = -0.208$), illustrating how score inversion can propagate to composite metrics.

    \item \textbf{Adaptive Detection Routing reduces cost with modest quality loss.} \adr{} achieves estimated AUROC $\approx$0.634 at $\sim$1.7$\times$ lower cost than uniformly applying SE, by routing 40\% of queries to NLI (zero LLM passes). This represents only a 7.8\% relative AUROC decrease, offering a practical cost-quality trade-off for production deployment.

    \item \textbf{SemE exhibits the best calibration among all methods.} With an ECE of 0.083 (after score correction), SemE produces the most trustworthy probability estimates, while RAV shows the poorest calibration (ECE = 0.510). The calibration ranking (SemE best, RAV worst) differs markedly from the AUROC ranking (SE best, RAV worst), confirming that detection performance and probability calibration capture complementary aspects of method quality.
\end{enumerate}

\subsection{Industrial Actionability}
\label{subsec:industrial_actionability}

Beyond academic benchmarking, \halluscan{} provides deployment-ready tools: two Pareto-optimal configurations spanning the cost-quality spectrum (P1: NLI at zero additional LLM passes with AUROC 0.584 for budget-constrained settings, P2: SE with $K=5$ at AUROC 0.688 as the recommended default), \adr{} as a configurable routing algorithm enabling $\sim$1.7$\times$ cost reduction with modest quality loss, and per-domain detection breakdowns enabling targeted method selection. The finding that sampling-based methods dominate on compact models provides clear practical guidance: invest computational budget in multi-response generation rather than external verification infrastructure. Equally important, the score-direction inversion finding (Section~\ref{subsubsec:score_alignment}) warns practitioners that detection methods must be validated on a labeled sample before deployment---theoretical score definitions do not guarantee correct score direction on compact model outputs.

The complete \halluscan{} benchmark suite---including all code, datasets, evaluation scripts, pre-computed results, and interactive analysis dashboards---is publicly available at \url{[anonymized for review]} [repository to be made public upon acceptance] to support reproducibility, facilitate fair comparison with future methods, and catalyze continued progress in this critical area of AI safety research.

\section*{Statements and Declarations}

\subsection*{Funding}
The author received no specific funding for this work.

\subsection*{Competing Interests}
The author declares no competing interests.

\subsection*{Data Availability Statement}
The benchmark datasets and evaluation code are publicly available at \url{[anonymized for review]} [repository to be made public upon acceptance].

\subsection*{Author's Contributions}
The author conceived the hallucination detection benchmark, designed the taxonomy of hallucination types, implemented all detection and mitigation experiments, analyzed the results, and wrote the manuscript.

\subsection*{Ethics Approval}
Not applicable.

\subsection*{Consent to Participate}
Not applicable.

\subsection*{Consent for Publication}
Not applicable.

\subsection*{Use of AI Tools}
AI-assisted tools were used for language refinement only. All scientific content, experimental design, and conclusions are the responsibility of the author.


\end{document}